\documentclass[10pt,twocolumn,letterpaper]{article}

\usepackage[pagenumbers]{cvpr} %

\usepackage[accsupp]{axessibility}  %
\usepackage{xcolor}
\usepackage{booktabs}
\usepackage{graphicx}
\usepackage{amsmath}
\usepackage{diagbox}
\usepackage{multirow}
\usepackage{multicol}
\usepackage{makecell}
\usepackage[export]{adjustbox}
\usepackage{tabularx}
\usepackage{longtable}
\usepackage{graphicx}
\usepackage{subcaption}
\usepackage{caption}
\usepackage{colortbl}
\usepackage{bm}
\usepackage{tcolorbox}
\usepackage{wrapfig}

\DeclareUnicodeCharacter{2194}{$\leftrightarrow$}
\definecolor{res_pred}{RGB}{128, 180, 102}
\definecolor{kl_reg}{RGB}{150, 116, 164}

\newcolumntype{Y}{>{\centering\arraybackslash}X}

\newcommand{\Ours}[0]{\textsc{MatLat}}

\newcommand{\matvae}[0]{\textsc{MatVAE}}

\newcommand{\receptacle}[1]{%
  \begingroup
  \setlength{\fboxsep}{2pt}%
  \colorbox{olive!15}{#1}%
  \endgroup
}
\newcommand{\targetobj}[1]{%
  \begingroup
  \setlength{\fboxsep}{2pt}%
  \colorbox{blue!15}{#1}%
  \endgroup
}

\definecolor{cvprblue}{rgb}{0.21,0.49,0.74}
\usepackage[pagebackref,breaklinks,colorlinks,allcolors=cvprblue]{hyperref}
\hypersetup{pdftitle={MatLat: Material Latent Space for PBR Texture Generation},pdfauthor={Kyeongmin Yeo, Yunhong Min, Jaihoon Kim, Minhyuk Sung}}

\def\paperID{} %
\def\confName{CVPR}
\def\confYear{2026}

\title{\Ours{}: Material Latent Space for PBR Texture Generation}

\author{Kyeongmin Yeo \quad Yunhong Min \quad Jaihoon Kim \quad Minhyuk Sung \\[0.2em]
KAIST\\
}

\begin{document}

\twocolumn[{%
\renewcommand\twocolumn[1][]{#1}%
\maketitle
\includegraphics[width=1.0\linewidth]{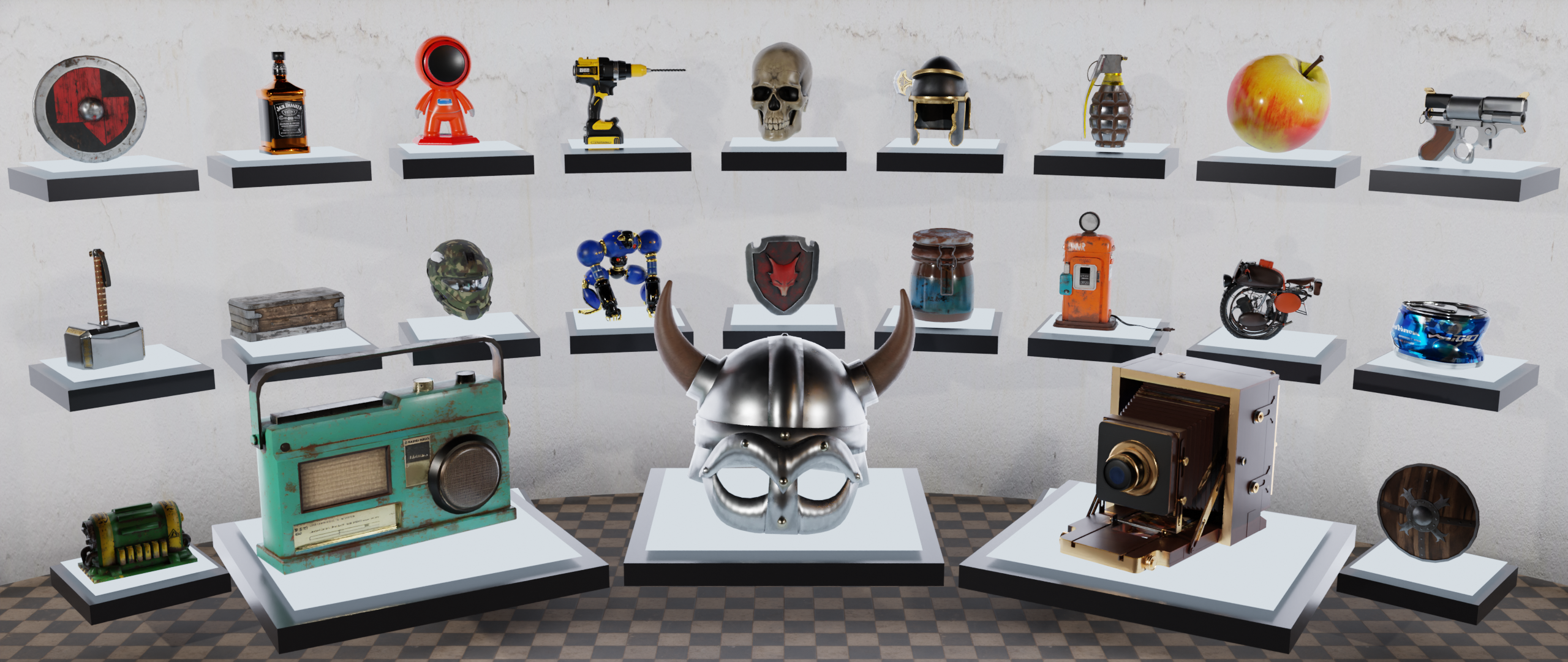}
\vspace{-1.8em}
\captionof{figure}{
\textbf{PBR Textures Generated by~\Ours{}.} 
Our method produces PBR textures that accurately represent rough materials (left), metallic surfaces (middle), and complex mixed materials (right). 
\vspace{0.7em}
}
\label{fig:teaser}
}]

\begin{abstract}
We propose a generative framework for producing high-quality PBR textures on a given 3D mesh. 
As large-scale PBR texture datasets are scarce, our approach focuses on effectively leveraging the embedding space and diffusion priors of pretrained latent image generative models while learning a material latent space,~\Ours{}, through targeted fine-tuning. Unlike prior methods that freeze the embedding network, which leads to distribution shifts when encoding additional PBR channels and hinders subsequent diffusion training, we fine-tune the pretrained VAE so that new material channels can be incorporated with minimal latent distribution deviation. We further show that correspondence-aware attention alone is insufficient for cross-view consistency unless the latent-to-image mapping preserves locality. To enforce this locality, we introduce a regularization in the VAE fine-tuning that crops latent patches, decodes them, and aligns the corresponding image regions to maintain strong pixel–latent spatial correspondence. 
Ablation studies and comparison with previous baselines demonstrate that our framework improves PBR texture fidelity and that each component is critical for achieving state-of-the-art performance. Our project page is available at \url{https://matlat-proj.github.io}.
\end{abstract}
    
\vspace{-1.3\baselineskip}
\section{Introduction}
\label{sec:intro}
\vspace{-0.25\baselineskip}

Generative modeling has recently driven remarkable progress in 3D asset creation. Modern methods can synthesize geometry~\cite{Jun:2023shap-e, Poole:2022dreamfusion, Chen:2023fantasia3d, Yi:2024gaussiandreamer}, textures and materials~\cite{Chen:2023text2tex,Richardson:2023texture, Kim:2024synctweedies, Yeo:2025stochsync}, motions and animations~\cite{Tevet:2023humanmotiondiffusionmodel, Hassan:2023synthesizing, Petrovich:2023tmr}, and even complete assets and scenes~\cite{Zhang:2024scenewiz3d,Höllein:2023text2room,Li:2024dreamscene}. Among these, a crucial requirement for production-ready assets is Physically Based Rendering (PBR) textures, which comprise albedo (diffuse color), roughness, and metallic channels. These channels enable physically accurate relighting under arbitrary illumination and are essential components of modern graphics pipelines. 

Recent works on PBR texture generation~\cite{Youwang:2024paintit,Zhang:2024dreammat,Huang:2024materialanything,He:2025materialmvp,Wang:2024boosting3d,Chen:2025meshgen} advance beyond baked appearances by predicting material maps that enable relightable assets.
However, progress remains constrained by the scarcity of large-scale, high-quality datasets with PBR channels, as existing public datasets~\cite{deitke2023objaverse, Deitke:2023objaversexl} provide only tens of thousands of such assets; even with recent additions such as TexVerse~\cite{zhang2025texverse}, their scale remains orders of magnitude smaller than that of image datasets.

A natural solution is to leverage pretrained image generative models with strong priors learned from large-scale RGB image datasets~\cite{Schuhmann:2022laion5b}. 
Techniques such as Score Distillation Sampling (SDS)~\cite{Poole:2022dreamfusion} and its variants~\cite{Wang:2023prolificdreamer, Lin:2023magic3d} have been explored for this purpose, but they often fail to produce high-fidelity PBR material images and frequently exhibit saturation artifacts. In practice, the most effective strategy has been to generate multi-view images with pretrained 2D models and then project them onto a mesh~\cite{Zhu:2024mcmat,Huang:2024materialanything,Wei:2025pbr3dgen,Zhu:2025muma,Fei:2025pacture,He:2025materialmvp,Chen:2025meshgen,Kim:2024synctweedies,Yeo:2025stochsync}. 
However, this multi-view generation approach still faces two major challenges for PBR texture synthesis. 
First, directly leveraging diffusion models trained on latent RGB images for PBR material generation is non-trivial, as PBR textures include additional channels (roughness and metallic). Simply mapping these channels to the pretrained encoder leads to a substantial domain gap.
Second, preserving multi-view consistency is challenging, and its failure can lead to blurring and visual artifacts in overlapping regions when the images are unprojected onto the mesh surface.

To this end, we introduce a novel framework that addresses both challenges by learning a \textbf{Mat}erial \textbf{Lat}ent space, referred to as~\Ours{}, which substantially improves the quality of generated PBR textures.
Specifically, we propose a two-stage pipeline: we first learn the material latent space by fine-tuning the pretrained VAE using PBR texture images, resulting in 
Material VAE (\matvae{}), and then fine-tune a diffusion model to generate multi-view material images in the adapted latent space.
Within this framework, we introduce the following two main components addressing the aforementioned challenges.

The first component is a latent-space adaptation module in~\matvae{} that extends the pretrained latent space to incorporate additional PBR channels—roughness and metallic—while preserving the pretrained prior.  
Most previous works~\cite{Zhu:2024mcmat,Huang:2024materialanything,Wei:2025pbr3dgen,Zhu:2025muma,Fei:2025pacture,He:2025materialmvp,Chen:2025meshgen} adopt a na\"ive approach that freezes the pretrained encoder and incorporates the additional PBR channels through zero-channel padding. 
While straightforward, this causes a severe distributional mismatch between the encoded PBR materials and the pretrained latent space, which leads to suboptimal performance during subsequent diffusion model fine-tuning. 
In contrast, our approach fine-tunes the pretrained encoder to effectively incorporate the additional channels of PBR material images while applying a distributional regularization that constrains deviations from the original latent distribution. 
This controlled adaptation effectively preserves the pretrained prior, stabilizes subsequent diffusion fine-tuning, and yields higher-fidelity PBR textures.
While similar latent-space extensions have been explored for transparent~\cite{Zhang:2024layerdiffuse} or RGB-D image synthesis~\cite{Krishnan:2025orchid}, this work is the \emph{first} to adapt this idea to PBR texture generation, and we further identify crucial components that make such adaptation effective in practice. 

The second component is locality regularization in \matvae{}, which ensures latent–image spatial alignment to improve multi-view consistency together with correspondence-aware attention (CAA)~\cite{Tang:2023mvdiffusion}.
Previous work on multi-view generation~\cite{Tang:2023mvdiffusion} has shown that CAA effectively enforces correspondence across views in the pretrained RGB latent space when the latent-image mapping preserves spatial locality. 
While such locality largely holds for the pretrained RGB encoder~\cite{He:2025materialmvp}, it often breaks for additional material channels, leading to multi-view inconsistent images. 
To address this, we introduce \emph{locality regularization}, applied during~\matvae{} fine-tuning, which explicitly enforces spatial alignment between latent tokens and image pixels. 
Specifically, we crop latent patches, decode them, and align the corresponding image regions with an $\ell_2$ reconstruction loss, encouraging pixel–latent spatial locality.

In our experiments, we evaluate the quality of the generated PBR textures against comprehensive baselines, including models trained from scratch, SDS-based optimization methods, and multi-view diffusion models. 
Quantitative results with comprehensive evaluation metrics show that \Ours{} outperforms baselines trained from scratch, which suffer from suboptimal quality due to limited PBR supervision, and SDS-based methods, which tend to produce over-saturated appearances and exhibit prohibitive runtimes. 
Notably, our method also outperforms multi-view diffusion baselines on most quantitative metrics, achieving new state-of-the-art performance. 
Additionally, we present systematic ablation studies that validate the effectiveness of each component of our framework, including the proposed latent-space adaptation in \matvae{}, correspondence-aware attention, and locality regularization. 

\vspace{-0.25\baselineskip}
\section{Related Work}
\label{sec:related_work}

\subsection{PBR Texture Generation}
Synthesizing PBR textures for 3D assets has emerged as an active area of research, driven by the growing integration of generative models into the 3D content creation pipeline. 
One line of research~\cite{Siddiqui:2024assetgen, Xiong:2025texgaussian} focuses on training models that directly generate 3D representations with associated material attributes from scratch. 
However, due to the limited availability and diversity of high-quality PBR texture data, these approaches exhibit suboptimal quality.

To address data scarcity, another class of approaches leverages pretrained diffusion models with expressive generative prior. 
Score Distillation Sampling (SDS)~\cite{Poole:2022dreamfusion} leverages a pretrained text-to-image diffusion model to update the parameters of 3D representations and has been adapted to PBR texture generation~\cite{Youwang:2024paintit, Deng:2024flashtex, Zhang:2024dreammat, Aliev:2025castex}. 
However, SDS tends to produce over-saturated colors with simplified patterns, degrading fine details, while its gradient-based updates are computationally expensive.

A promising direction is to generate multi-view PBR material images (\eg, albedo, metallic, roughness) with diffusion models and then project them onto 3D meshes to obtain the texture maps. 
This approach exploits the expressive prior of pretrained diffusion models while enforcing multi-view consistency via cross-view attention or correspondence modules~\cite{Zhu:2024mcmat,Huang:2024materialanything,Wei:2025pbr3dgen,Zhu:2025muma,Fei:2025pacture,He:2025materialmvp,Chen:2025meshgen}. 
From this perspective, we adopt a multi-view image generation framework and present \Ours{} which addresses two key challenges: (i) leveraging pretrained diffusion priors and (ii) ensuring multi-view consistency.

In the following subsections, we review previous works addressing these components, including approaches not limited to PBR material generation.

\subsection{Leveraging Pretrained Diffusion Priors}

Leveraging the generative prior of large-scale pretrained diffusion models is crucial for generating data in modalities with limited availability (\eg, normals, depth, alpha)~\cite{Zeng:2024rgbx, Li:2025idarb, Fu:2024geowizard, Yu:2025unicon, Kwon:2025jointdit}. 
A key challenge, however, lies in extending a diffusion model trained on latent RGB images to target modalities whose dimensionality differs due to the additional channels.
A straightforward approach~\cite{He:2025materialmvp} processes the target modality image by splitting and padding its channels to form a three-channel tensor, feeding it to a frozen RGB-trained encoder, and then fine-tuning the diffusion model on the resulting latents. 
However, the encoded latent distribution generally exhibits a significant domain gap from the pretrained latent space since the target-modality channels (\eg, roughness and metallic) are fed into an RGB encoder. 
This mismatch prevents leveraging the pretrained diffusion prior and makes downstream adaptation inefficient.
To address this, LayerDiffuse~\cite{Zhang:2024layerdiffuse} and Orchid~\cite{Krishnan:2025orchid} adapt the VAE to align the latent representations of new modalities with the pretrained RGB latent space, reducing distributional discrepancies and improving the stability of subsequent diffusion fine-tuning.
Inspired by this, in Sec.~\ref{subsec:mcvae}, we also propose to fine-tune the pretrained VAE while identifying the best practice that achieves superior performance compared to previous methods.

\subsection{Multi-View-Consistent Generation}

Beyond leveraging pretrained diffusion priors, maintaining multi-view consistency is crucial for high-quality PBR textures, as view misalignments during projection onto the mesh blur details in overlapping regions. 
Previous works address this challenge through dense multi-view attention~\cite{Shi:2024mvdream, Wang:2023imagedream, Long:2023wonder3d, Gao:2024cat3d, He:2025materialmvp}.
While this design enables information sharing across views, it does not explicitly incorporate geometric information and often yields suboptimal view consistency.
A more effective approach is correspondence-aware attention (CAA)~\cite{Tang:2023mvdiffusion}, which leverages pixel-wise correspondence to enable geometry-guided feature sharing in the latent space. 
By restricting attention computation to geometrically corresponding regions, CAA improves cross-view consistency. 
However, CAA is most effective in the pretrained RGB latent space when latent–image spatial alignment is well preserved~\cite{Tang:2023mvdiffusion, Kant:2024spad, Li:2024era3d, Huang:2025mvadapter}. 
To improve it, in Sec.~\ref{subsec:mvdit}, we introduce \emph{locality regularization}, applied during encoder fine-tuning to enforce spatial alignment between \matvae{} latents and PBR images. 

\vspace{-0.2\baselineskip}
\section{\Ours{}: Material Latent Space} 
\label{sec:method}
\vspace{-0.2\baselineskip}
Our goal is to build a generative model that produces high-quality PBR textures for given meshes.
In particular, we propose a model that leverages the priors of pretrained image diffusion models by generating PBR textures as multi-view images and projecting them onto the mesh. 
Specifically, let $\mathcal{M}$ be a 3D mesh surface, and let $y$ be a text prompt specifying the target appearance. 
We denote a set of $N$ camera viewpoints $\{c_i\}_{i=1}^N$, where each viewpoint $c_i$ is defined by its rotation $\mathbf{R}_i\in\mathrm{SO}(3)$ and translation $\mathbf{t}_i\in\mathbb{R}^3$. 
These parameters define an orthographic projection $\Pi_i$ which maps the 3D surface $\mathcal{M}$ onto the discrete $H \times W$ pixel grid $\Omega=\{1,\dots,H\}\times\{1,\dots,W\}$. 
Then, for each view $c_i$, our objective is to produce a five-channel PBR material image $\mathbf{x}_i$ encoding albedo $\mathbf{a}_i \in \mathbb{R}^{H \times W \times 3}$, roughness $\mathbf{r}_i \in \mathbb{R}^{H \times W \times 1}$, and metallic $\mathbf{m}_i \in \mathbb{R}^{H \times W \times 1}$. 

The key component of our framework is the representation of PBR material images, which we refer to as~\Ours{}. 
We design a latent PBR material space that jointly encodes all PBR channels while preserving the dimensionality of the original latent space.
To this end, we introduce \matvae{}, which maps PBR material images into a latent space aligned with the pretrained latent distribution via (i) residual prediction and (ii) KL regularization (Sec.~\ref{subsec:mcvae}). 
This alignment improves convergence during subsequent diffusion-model adaptation.
Meanwhile, to achieve multi-view consistency, we introduce correspondence-aware attention (CAA), integrated into the diffusion model. 
More importantly, we propose a novel locality regularization applied during \matvae{} training, which preserves latent–image spatial alignment and improves view consistency of the generated PBR material images (Sec.~\ref{subsec:mvdit}).

\subsection{Material VAE (\matvae{})}
\label{subsec:mcvae}
We present \matvae{} for encoding PBR material images while preserving the pretrained latent distribution. 
Unless stated otherwise, material images are rendered from a single viewpoint, and we omit the view subscript for simplicity. 
Before introducing our \matvae{}, we first discuss Frozen VAE, a widely used approach for modeling material latents in PBR texture generation.

\subsubsection{Prior Work: Frozen VAE~\cite{He:2025materialmvp}}
\begin{figure}[h!]
    \centering
    \scriptsize
    \vspace{-0.5\baselineskip}
    \includegraphics[width=\linewidth]{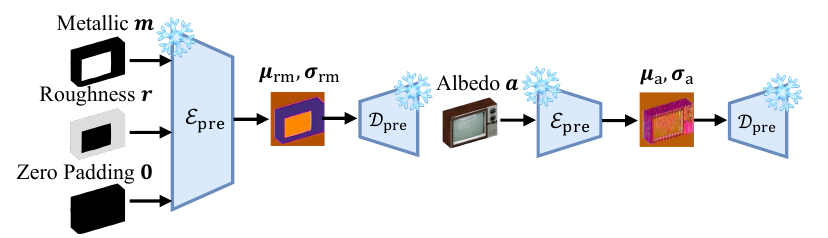}
    \vspace{-1.5\baselineskip}
    \caption{
        \textbf{Overview of Frozen VAE~\cite{He:2025materialmvp}.}
        Zero-padded roughness–metallic maps and albedo image are encoded by the frozen VAE to produce $(\boldsymbol{\mu}_{\text{rm}}, \boldsymbol{\sigma}_{\text{rm}})$ and $(\boldsymbol{\mu}_{\text{a}}, \boldsymbol{\sigma}_{\text{a}})$, respectively. 
    }
    \label{fig:frozen_vae}
    \vspace{-\baselineskip}
\end{figure}

Leveraging the generative prior of pretrained diffusion models is critical for producing high-quality PBR material images. 
To this end, most previous works~\cite{Zhu:2024mcmat,Huang:2024materialanything,Wei:2025pbr3dgen,Zhu:2025muma,Fei:2025pacture,Chen:2025meshgen}, including MaterialMVP~\cite{He:2025materialmvp}, adopt a na\"ive approach, which we term Frozen VAE, where the pretrained VAE is frozen and its encoder is reused to obtain latents for five-channel PBR input images.
To match the three-channel input format of the pretrained VAE, the methods separately encode the albedo from the roughness and metallic channels, with the roughness and metallic concatenated with zero-padding before encoding. 
An overview of Frozen VAE design is shown in Fig.~\ref{fig:frozen_vae}, where $(\mathcal{E}_\text{pre}, \mathcal{D}_\text{pre})$ denote the pretrained encoder and decoder, respectively. 
Formally, given a PBR material image $\mathbf{x} = [\mathbf{a}, \mathbf{r}, \mathbf{m}]$, 
\begin{align}
    \nonumber 
    &(\boldsymbol{\mu}_\text{rm}, \boldsymbol{\sigma}_\text{rm}) = \mathcal{E}_\text{pre}([\mathbf{r}, \mathbf{m}, \mathbf{0}]), 
    \hspace{1mm} 
    (\boldsymbol{\mu}_\text{a}, \boldsymbol{\sigma}_\text{a}) = \mathcal{E}_\text{pre}(\mathbf{a}), 
    \text{ and} \\
    \nonumber 
    & q(\mathbf{z}_{\text{rm}} | \mathbf{r}, \mathbf{m}) = \mathcal{N}(\boldsymbol{\mu}_\text{rm}, \boldsymbol{\sigma}_\text{rm}^2), 
    \hspace{1mm}
    q(\mathbf{z}_{\text{a}} | \mathbf{a}) = \mathcal{N} (\boldsymbol{\mu}_\text{a}, \boldsymbol{\sigma}_\text{a}^2),
\end{align}
where $\bm{0} \in \mathbb{R}^{H \times W \times 1}$ is the zero-padded channel. 
Here, $\boldsymbol{\mu}_{\{\text{rm}, \text{a}\}}, \boldsymbol{\sigma}_{\{\text{rm}, \text{a}\}}$ denote the mean and standard deviation for the roughness/metallic and albedo latents, respectively. 
The diffusion model is then fine-tuned using the concatenated latents $[\mathbf{z}_\text{rm}, \mathbf{z}_\text{a}]$.

While $\mathbf{z}_\text{a}$ aligns closely with the pretrained latent space due to the semantic similarity of albedo to RGB images~\cite{He:2025materialmvp}, $\mathbf{z}_\text{rm}$ introduces a substantial domain gap, as roughness and metallic images exhibit distinct visual characteristics. 
This yields out-of-distribution latents that deviate from the representation space of pretrained diffusion models and, consequently, leads to suboptimal generation quality.
We provide additional empirical evidence for this latent mismatch in the appendix (Sec.~\ref{subsec:latent_mismatch}).
Moreover, generating two separate latent codes doubles the inference cost of the diffusion model.

\begin{figure*}
    \centering
    \scriptsize
    \includegraphics[width=\linewidth]{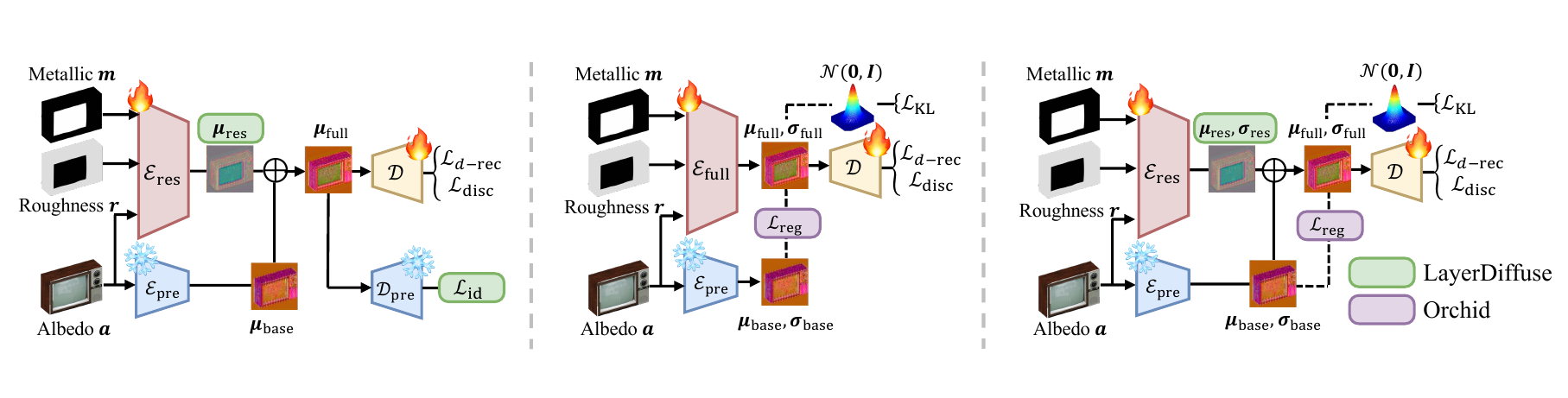}

    \vspace{-0.2\baselineskip}
    \begin{minipage}{0.31\linewidth}
    \centering
    \vspace{-2.6\baselineskip} 
    \small (a) Res. Pred. + $\mathcal{L}_{\text{id}}$ (LayerDiffuse~\cite{Zhang:2024layerdiffuse})
    \end{minipage}%
    \begin{minipage}{0.34\linewidth}
    \centering
    \vspace{-2.6\baselineskip} 
    \hspace{0.3em}\small (b) Direct Pred. + $\mathcal{L}_{\text{reg}}$ (Orchid~\cite{Krishnan:2025orchid})
    \end{minipage}%
    \begin{minipage}{0.34\linewidth}
    \centering
    \vspace{-2.6\baselineskip} 
    \hspace{-1em}\small (c) Res. Pred. + $\mathcal{L}_{\text{reg}}$ ({\matvae{}})
    \end{minipage}
    \vspace{-1.1\baselineskip}
    \caption{
        \textbf{
            {Comparison of PBR Material Encoder Schemes.}
        }
        (a) LayerDiffuse~\cite{Zhang:2024layerdiffuse} uses residual prediction but only predicts the latent mean and enforces identity consistency via $\mathcal{L}_{\text{id}}$. 
        (b) Orchid~\cite{Krishnan:2025orchid} uses direct prediction to output the full latent parameters $(\boldsymbol{\mu}_{\text{full}}, \boldsymbol{\sigma}_{\text{full}})$ and regularizes them with $\mathcal{L}_{\text{reg}}$. 
        (c) \matvae{} (ours) synergistically integrates \textcolor{res_pred}{residual prediction} and \textcolor{kl_reg}{KL regularization}, enabling effective incorporation of PBR material images while preserving the pretrained latent space.  
    }
    \label{fig:encoder_design}
    \vspace{-1.3\baselineskip}
\end{figure*}

\subsubsection{Residual Prediction and KL Regularization}
To address this, we propose \matvae{}, which adapts the pretrained VAE to encode PBR channels into a single latent code while preserving the pretrained diffusion prior.
Motivated by the semantic similarity between albedo and RGB images in Frozen VAE, we reuse $\mathcal{E}_{\text{pre}}$ to encode the albedo, \ie, $q(\mathbf{z}_\text{base} | \mathbf{a}) = \mathcal{N}\big(\boldsymbol{\mu}_{\text{base}}, \boldsymbol{\sigma}_{\text{base}}^{2} \big)$, which serves as the base latent distribution.  
On the other hand, we introduce a learnable \emph{residual encoder} $\mathcal{E}_{\text{res}}$ 
to \emph{inject} the roughness and metallic information. 
Specifically, as shown in Fig.~\ref{fig:encoder_design}(c), the encoder predicts residual parameters $(\boldsymbol{\mu}_\text{res}, \boldsymbol{\sigma}_\text{res}) = \mathcal{E}_{\text{res}}(\mathbf{x})$, 
which adjust the base latent distribution: 
\vspace{-0.4\baselineskip}
{\normalsize
\begin{align}
\label{eq:res_pred}
\mathbf{z} \sim \mathcal{N} \left(\boldsymbol{\mu}_{\text{base}} + \boldsymbol{\mu}_{\text{res}}, \boldsymbol{\sigma}^2_{\text{base}} \odot \boldsymbol{\sigma}^2_{\text{res}} \right) = q(\mathbf{z} | \mathbf{x}),
\end{align}
}%
\noindent
where $\odot$ denotes element-wise multiplication. 
For notation convenience, we define $(\boldsymbol{\mu}_{\text{full}}, \boldsymbol{\sigma}_{\text{full}})= (\boldsymbol{\mu}_{\text{base}} + \boldsymbol{\mu}_{\text{res}}, \boldsymbol{\sigma}_{\text{base}} \odot \boldsymbol{\sigma}_{\text{res}})$, denoting the mean and standard deviation of latent $\mathbf{z}$, respectively.  
We refer to the formulation above as residual prediction.

With a slight abuse of notation, let $\mathcal{E}=(\mathcal{E}_{\text{pre}}, \mathcal{E}_{\text{res}})$ denote the combination of the pretrained and residual encoders, and write $\mathcal{E}(\mathbf{x})$ for a latent sample $\mathbf{z}$ drawn from the Gaussian distribution as in Eq.~\ref{eq:res_pred}. 
During training, a learnable decoder $\mathcal{D}$ maps the latent code $\mathbf{z}$ back to the five-channel PBR material image, \ie, $\hat{\mathbf{x}}=\mathcal{D}(\mathbf{z})$. 
Additionally, the pretrained encoder is kept frozen, and $\mathcal{E}, \mathcal{D}$ are optimized using the following objectives:
\begin{align}
    \label{eq:loss_rec}
    \mathcal{L}_{d\text{-rec}}
    &= \lambda_{d\text{-rec}} \cdot d \left( \mathbf{x}, \hspace{1mm} \mathcal{D} (\mathcal{E} (\mathbf{x})) \right), \\
    \label{eq:loss_kl}
    \mathcal{L}_{\text{KL}}
    &= \lambda_\text{KL} \cdot \mathrm{KL}(q(\mathbf{z} | \mathbf{x}) \| \mathcal{N}(\mathbf{0}, \mathbf{\textit{I}})), \\
    \label{eq:loss_disc}
    \mathcal{L}_{\text{disc}}
    &= \lambda_{\text{disc}} \cdot \mathcal{F}(\mathcal{D} (\mathcal{E}(\mathbf{x}))),
\end{align}
where $\lambda_{(\cdot)}$ are the weights for each loss term, $\mathcal{F}$ denotes the discriminator used in the adversarial loss~\cite{Goodfellow:2014gan}, and $d(\cdot,\cdot) \in \{\ell_2, \text{LPIPS}\}$ specifies the reconstruction distance metric.

While these objectives ensure faithful reconstruction and effective compression of PBR material images, the learned representation space can deviate from the pretrained latent distribution. 
To address this, we introduce a regularizer that penalizes the divergence between the learned latent distribution and that of the pretrained model:
\begin{align}
    \label{eq:loss_distill}
    \mathcal{L}_{\text{reg}}
    &= \lambda_{\text{reg}} \cdot \mathrm{KL} 
    ( 
    \underbrace{q(\mathbf{z} | \mathbf{x})}_{\text{Learned}} \| \underbrace{q(\mathbf{z}_\text{base} | \mathbf{a})}_{\text{Pretrained}}
    ). 
\end{align}
The schematic of the KL regularization is presented in Fig.~\ref{fig:encoder_design}(b). 
This regularizer constrains the learned latent distribution to remain close to the pretrained latent space, ensuring compatibility with the diffusion prior.

Putting these together, we introduce our Material VAE (\matvae{}) which combines i) \textcolor{res_pred}{residual prediction} (Eq.~\ref{eq:res_pred}) and ii) \textcolor{kl_reg}{KL regularization} (Eq.~\ref{eq:loss_distill}).
The schematic is shown in Fig.~\ref{fig:encoder_design}(c), with each component highlighted. 

\vspace{-\baselineskip}
\paragraph{Connection to LayerDiffuse~\cite{Zhang:2024layerdiffuse}.} 
Residual prediction of VAE has also been explored for transparent image generation~\cite{Zhang:2024layerdiffuse}. 
However, LayerDiffuse only predicts the latent mean, effectively treating the encoder as deterministic rather than stochastic, \ie, $q(\mathbf{z} | \mathbf{x}) = \delta (\mathbf{z} - \boldsymbol{\mu}_{\text{full}})$, which may disrupt the smooth latent representation of the pretrained VAE.
Although an identity-consistency loss $\mathcal{L}_{\text{id}} = d \left(\mathbf{a}, \mathcal{D}_\text{pre} (\mathcal{E}(\mathbf{x}))\right)$ is employed, it does \emph{not} enforce distribution-level alignment with the pretrained latent space. 
In contrast, we predict both the mean and standard deviation of the residual latent and introduce a KL regularizer that enforces distributional alignment.

\vspace{-0.7\baselineskip}
\paragraph{Connection to Orchid~\cite{Krishnan:2025orchid}.}
We note that KL regularization has also been explored in Orchid~\cite{Krishnan:2025orchid} for depth and normal map generation. 
However, Orchid directly predicts $(\boldsymbol{\mu}_\text{full}, \boldsymbol{\sigma}_\text{full})$, while our \matvae{} adopts a residual prediction, in which the final convolution layer of $\mathcal{E}_\text{res}$ is initialized with zero weights, yielding zero residual predictions at the start of training and thus exactly reproducing the pretrained latent distribution. 
As observed in previous works~\cite{zhang2023adding}, this initialization strategy stabilizes optimization and allows the latent space to be gradually adapted to the new modality. 

\vspace{-0.7\baselineskip}
\paragraph{Our Contributions.}
To the best of our knowledge, this is the first work to introduce effective fine-tuning of latent embeddings for PBR texture generation. In our experiments (Sec.~\ref{sec:experiments}), we further demonstrate that the implementation details which differ from those of previous approaches play an important role in achieving the best performance.

\vspace{0.1\baselineskip}
\subsection{Multi-View Consistent PBR Texture}
\label{subsec:mvdit}
In this section, we present two key components that together enforce multi-view consistency in PBR material generation.

\subsubsection{Correspondence-Aware Attention}
In addition to learning a valid material representation, PBR material images must satisfy multi-view consistency.
However, the dense multi-view attention used in previous works~\cite{He:2025materialmvp} does not leverage geometric priors, leading to inefficient computation and suboptimal results.

Formally, let $\mathbf{Z} \in \mathbb{R}^{Nhw \times d}$ denote the latent tokens from $N$ viewpoints, each with spatial resolution $h \times w$, where $d$ is the channel dimension. 
Then, a dense multi-view attention computes: 
{\small
\begin{align}
    \nonumber
    \mathrm{softmax}\big(\mathbf{Q}\mathbf{K}^\top / \sqrt{d}\big)\mathbf{V}, 
    \hspace{0.5mm}  \mathbf{Q} = \mathbf{W}_Q\mathbf{Z}, 
    \hspace{0.5mm} \mathbf{K} = \mathbf{W}_K\mathbf{Z}, 
    \hspace{0.5mm} \mathbf{V} = \mathbf{W}_V\mathbf{Z},
\end{align}
}%
\noindent
where $\mathbf{W}_{\{Q, K, V\}}$ denote the query, key, and value projections, respectively. 
Note that each token attends to all others across views without geometric or visibility priors, forcing the network to learn correspondences implicitly. 
Consequently, the model converges slowly and produces view-inconsistent results~\cite{He:2025materialmvp}.

\begin{figure*}[t!]
    \centering
    \scriptsize
    \includegraphics[width=\linewidth]{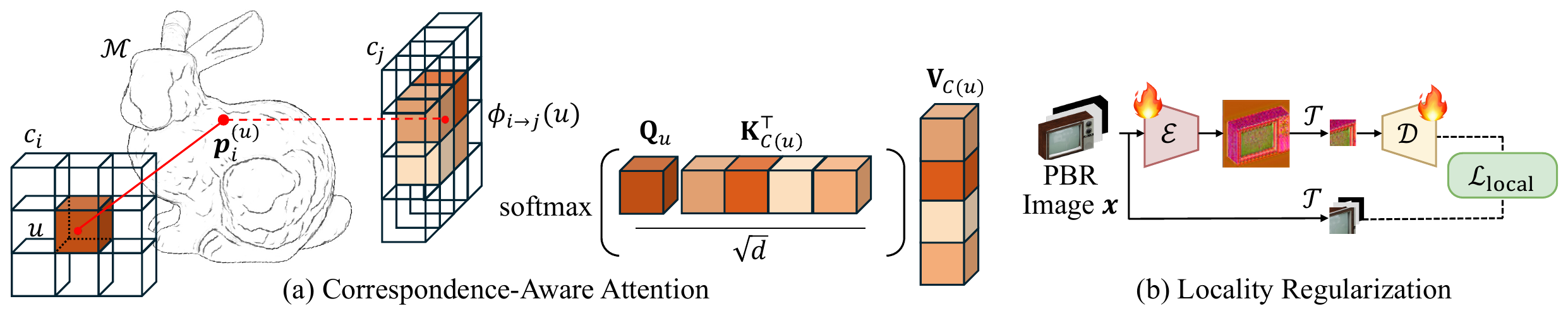}
    \vspace{-2\baselineskip}
    \caption{
    \textbf{Illustration of Correspondence-Aware Attention (CAA) and Locality Regularization.}
    (a) CAA restricts attention to geometrically corresponding tokens across views. 
    (b) Locality regularizer enforces patch-wise reconstruction such that image pixels are decoded from spatially aligned latent tokens. 
    Together, CAA and locality regularization enable multi-view consistent PBR texture generation.
    }
  \label{fig:caa_patch_rec}
  \vspace{-1.5\baselineskip}
\end{figure*}

To address the limitation of dense multi-view attention, correspondence-aware attention (CAA)~\cite{Tang:2023mvdiffusion} incorporates \emph{explicit} point correspondences into the attention computation. 
In Fig.~\ref{fig:caa_patch_rec}(a), we illustrate the CAA operation. 
Formally, for a pixel \(u \in \Omega\) in view $c_i$, let the observed surface point $\mathbf{p}_i^{(u)} = \Pi_i^{-1}(u) \in \mathcal{M}$. 
Additionally, for every view \(j \neq i\), we define the correspondence pixel as the nearest-neighbor discretization of the projection $\phi_{i \rightarrow j}(u) = \Pi_j(\mathbf{p}_i^{(u)})$ if the surface point is visible in view $c_j$. 
Then, the set of valid correspondences for $u$ across the remaining $N-1$ views is $\mathcal{C}(u) = \big\{  \phi_{i \rightarrow j}(u) \big| j \neq i \big\}$. 
In practice, correspondences are implemented as a \(K \times K\) local window centered at \(\phi_{i \rightarrow j}(u)\). 
CAA is then computed by restricting computation to the correspondence set $\mathcal{C}(u)$:
\begin{align}
    \label{eq:caa}
    \mathrm{softmax}\big(\mathbf{Q}_u \mathbf{K}_{\mathcal{C}(u)}^\top / \sqrt{d} \big)\ \mathbf{V}_{\mathcal{C}(u)},
\end{align}
where $\mathbf{Q}_u$ is the query associated with token $u$, $\mathbf{K}_{\mathcal{C}(u)}$ and $\mathbf{V}_{\mathcal{C}(u)}$ are the keys and values of its correspondence set. 
Providing explicit correspondences in CAA strengthens cross-view alignment, yielding multi-view consistent PBR material images.

\subsubsection{Locality Regularization}
\label{subsubsec:locality_reg}
While CAA promotes cross-view feature exchange in latent space, our objective is to enforce multi-view consistency in the decoded pixel-space images. 
Hence, using CAA alone to achieve view consistency in pixel space requires spatial locality between latent tokens and their corresponding image pixels, \ie, each image pixel is primarily decoded from its spatially local latent tokens. 
While this assumption holds for the pretrained encoder~\cite{Tang:2023mvdiffusion, Kant:2024spad, Li:2024era3d, Huang:2025mvadapter}, \matvae{}, which is fine-tuned to encode additional PBR channels under prior preservation regularization, may not preserve the latent-image locality. 
Consequently, applying CAA to \matvae{} latents results in information exchange between geometrically unrelated tokens, degrading multi-view consistency.

To ensure latent–image locality, we introduce a locality regularizer that enforces patch-wise reconstruction during \matvae{} fine-tuning:
\begin{align}
    \label{eq:loss_equi_rec}
    \mathcal{L}_{\text{local}}
    = \lambda_{\text{local}} \cdot
    d \left(
    \mathcal{T} (\mathbf{x}), 
    \mathcal{D} (\mathcal{T} (\mathcal{E} (\mathbf{x})))
    \right),
\end{align}
where $d(\cdot,\cdot)=\ell_2$ is a distance metric, and $\mathcal{T}$ is a random crop operator. 
We present a visualization of the regularizer in Fig.~\ref{fig:caa_patch_rec}(b). 
This regularization enforces spatial locality by ensuring each image pixel is reconstructed mainly from aligned latent tokens.

Accordingly, we replace the reconstruction loss in Eq.~\ref{eq:loss_rec} with the locality regularization, yielding the following \matvae{} training objective:
\vspace{-0.2\baselineskip}
\begin{align}
    \label{eq:matvae_loss}
    \mathcal{L}_{\matvae{}} = 
    \mathcal{L}_{\text{local}} +
    \mathcal{L}_{\text{KL}} +
    \mathcal{L}_{\text{disc}} +
    \mathcal{L}_{\text{reg}}. 
\end{align}
\vspace{-0.2\baselineskip}
In Sec.~\ref{sec:experiments}, we validate that CAA and locality regularizer synergistically enhance multi-view consistency.

\subsection{\Ours{} Training Objective}
\label{subsec:loss}
We fine-tune a latent diffusion (velocity) model~\cite{esser2024scaling}, denoted as $u_{\theta}$, on multi-view PBR latents
$\mathbf{Z}$ obtained by rendering the mesh from $N$ viewpoints and encoding each view with \matvae{}. 
Additionally, we insert CAA modules into the attention blocks of the diffusion model, in addition to the original attention layers.
Following Conditional Flow Matching (CFM)~\cite{Lipman:2023flowmatching}, we define a linear path between a data latent $\mathbf{Z}$ and Gaussian noise
$\boldsymbol{\epsilon} \sim \mathcal{N}(\mathbf{0},\mathbf{I})$:
$\mathbf{Z}_{t}=(1-t)\mathbf{Z} + t \boldsymbol{\epsilon}$ for $t\in[0,1]$. 
Our \Ours{} is then trained using the CFM objective:
\begin{equation}
\label{eq:matlat_loss}
\mathcal{L}_{\text{\Ours{}}}
=
\mathbb{E}_{t, \mathbf{Z}, \boldsymbol{\epsilon}}
\left\|
u_{\theta} \big(\mathbf{Z}_{t}, t, y \big) - \mathbf{u}_{t}
\right\|^{2},
\end{equation}
where $\mathbf{u}_{t} = \boldsymbol{\epsilon} - \mathbf{Z}$ is the instantaneous velocity, and $y$ is the conditioning text caption.

\vspace{-0.5\baselineskip}
\section{Experiments}
\vspace{-0.25\baselineskip}
\label{sec:experiments}

In this section, we present experimental results on PBR texture generation. 
We first describe the experimental setup (Sec.~\ref{subsec:exp_setup}), followed by ablation studies that validate the effectiveness of each component (Sec.~\ref{subsec:ablation}). 
Finally, we report quantitative comparisons against external baselines (Sec.~\ref{subsec:external_comp}). 
In the appendix, we present (i) technical discussions, (ii) implementation and experiment details, and (iii) additional qualitative results.

\vspace{-0.2\baselineskip}
\subsection{Experiment Setup}
\label{subsec:exp_setup}
We fine-tune \Ours{} from the pretrained VAE and diffusion model of \textsc{Stable Diffusion 3.5-medium}, using $40{,}723$ PBR-textured assets from Objaverse-XL~\cite{Deitke:2023objaversexl} paired with captions from Bootstrap3D~\cite{Sun:2025bootstrap3d} and Cap3D~\cite{Luo:2023cap3d} as training data.
During training, we render $26$ images from fixed viewpoints for each $3$D mesh. 

For generation, we use six canonical views (front, back, left, right, top, and bottom) following previous works~\cite{Huang:2024materialanything, He:2025materialmvp}. 
For evaluation, we follow PacTure~\cite{Fei:2025pacture} and use $128$ meshes not included in the training set, using $N=20$ camera views per mesh: eight views uniformly spaced in azimuth at $0^\circ$ elevation, six at $30^\circ$, four at $45^\circ$, and two polar views at $\pm 90^\circ$.
For each view, we render (i) albedo, roughness, and metallic images and (ii) a shaded RGB image under randomly sampled HDR environment maps. 
The reference images are rendered using the same set of $20$ camera views and the identical HDR environment maps with the ground-truth PBR textures. 

\begin{table*}[t]
  \centering
  \small
  \newcolumntype{x}{>{\centering\arraybackslash}m{0.01\textwidth}}
  \caption{\textbf{Quantitative Results of Ablation Study.} Best scores are \textbf{bold}, and the runner-up scores are \underline{underlined}.}
  \label{tab:ablation}
  \vspace{-0.5\baselineskip}
  \resizebox{\textwidth}{!}{
  \begin{tabular}{x l ccc ccc c c c c}
  \toprule
    \multirow{2}{*}{Id} &
    \multirow{2}{*}{\textbf{Methods}} &
    \multicolumn{3}{c}{\textbf{Shaded}} &
    \multicolumn{3}{c}{\textbf{Albedo}} &
    \textbf{Rough.} &
    \textbf{Metal.} &
    \multirow{2}{*}{\textbf{c-PSNR} $\uparrow$} &
    \multirow{2}{*}{\textbf{Time} $\downarrow$} \\

    \cmidrule(lr){3-5}\cmidrule(lr){6-8}\cmidrule(lr){9-9}\cmidrule(lr){10-10}
    & & 
    $\mathrm{FID}_{\text{CLIP}}\downarrow$ &
    $\mathrm{KID}\downarrow$ &
    $\mathrm{CLIP}\uparrow$ &
    $\mathrm{FID}_{\text{CLIP}}\downarrow$ &
    $\mathrm{KID}\downarrow$ &
    $\mathrm{CLIP}\uparrow$ &
    
    $\mathrm{RMSE}\downarrow$ &
    
    $\mathrm{RMSE}\downarrow$ &
    &\\
    \midrule

1 & Frozen VAE~\cite{He:2025materialmvp}
& 3.419
& \textbf{1.052}
& \underline{0.318}
& 4.926
& 1.560
& \textbf{0.315}
& 0.193
& 0.251
& 19.869
& 111s \\

\midrule

2 & \textbf{Res. Pred. + $\mathcal{L}_{\text{reg}}$ (Ours)}
& \textbf{3.083}
& 1.327
& \underline{0.318}
& \textbf{4.599}
& 1.574
& \underline{0.314}
& 0.158
& \underline{0.134}
& \textbf{21.934}
& 34s \\

3 & \hspace{1mm} $\llcorner$ Res. Pred. + $\mathcal{L}_{\text{id}}$~\cite{Zhang:2024layerdiffuse}
& 3.210
& 1.090
& 0.317
& 4.871
& 1.525
& 0.312
& 0.165
& \textbf{0.128}
& 20.977
& 34s \\

4 & \hspace{1mm} $\llcorner$ Direct Pred. + $\mathcal{L}_{\text{reg}}$~\cite{Krishnan:2025orchid}
& 3.192
& 1.146
& \textbf{0.319}
& 4.768
& \underline{1.472}
& 0.312
& 0.161
& 0.170
& \underline{21.468}
& 34s \\

\midrule

5 & \hspace{1mm} $\llcorner$ w/o $\mathcal{L}_\text{local}$
& 3.419
& 1.611
& 0.316
& 5.873
& 4.122
& 0.312
& \textbf{0.154}
& 0.182
& 19.437
& 34s \\

6 & \hspace{1mm} $\llcorner$ w/o CAA
& \underline{3.110}
& \underline{1.088}
& 0.317
& \underline{4.732}
& \textbf{1.388}
& 0.312
& \underline{0.155}
& 0.151
& 18.687
& 38s \\

\bottomrule
  \end{tabular}
  }
\vspace{-0.75\baselineskip}
\end{table*}

\begin{figure}[t]
  \scriptsize
  \centering
  \renewcommand{\arraystretch}{1.0}
  \setlength{\tabcolsep}{4pt}
  \begin{tabularx}{\linewidth}{YYYY}
    \makecell{Frozen VAE~\cite{He:2025materialmvp}} &
    \makecell{Res. Pred.\\+ $\mathcal{L}_{\text{id}}$~\cite{Zhang:2024layerdiffuse}} &
    \makecell{Direct Pred.\\+ $\mathcal{L}_{\text{reg}}$~\cite{Krishnan:2025orchid}} &
    \textbf{\makecell{Res. Pred.\\+ $\mathcal{L}_{\text{reg}}$ (\textbf{Ours})}} \\

    \includegraphics[width=0.115\textwidth,valign=m]{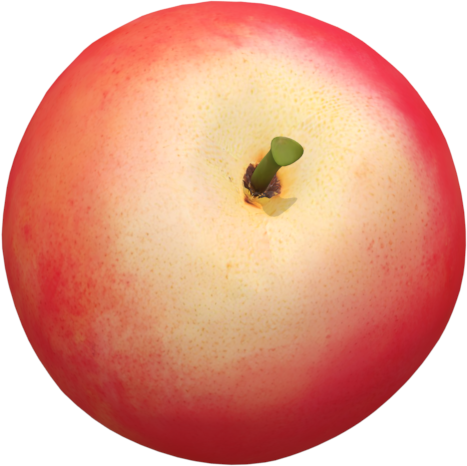} &
    \includegraphics[width=0.115\textwidth,valign=m]{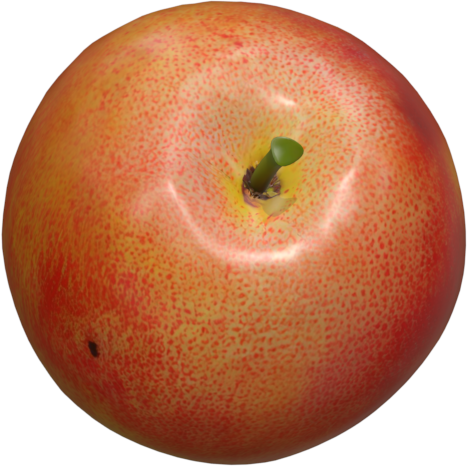} &
    \includegraphics[width=0.115\textwidth,valign=m]{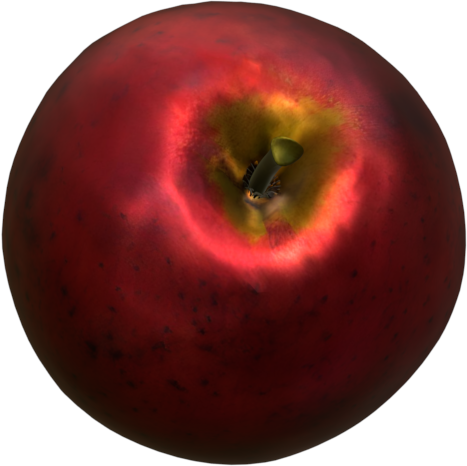} &
    \includegraphics[width=0.115\textwidth,valign=m]{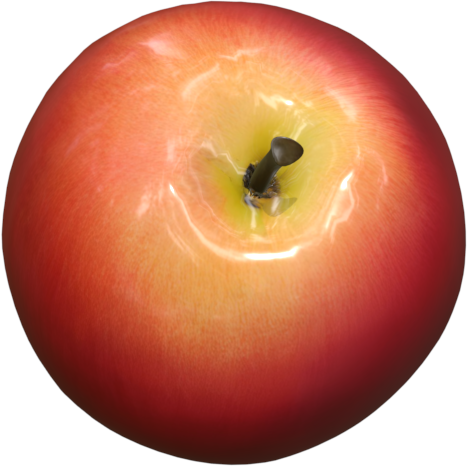}
    \\
    \multicolumn{4}{c}{\textit{\makecell{``A realistic apple with \receptacle{glossy red skin} mottled with \targetobj{yellow highlights}.''}}}\\
    \includegraphics[width=0.115\textwidth,valign=m]{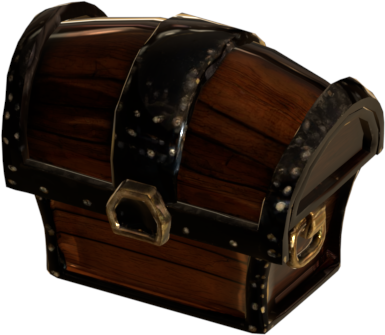} &
    \includegraphics[width=0.115\textwidth,valign=m]{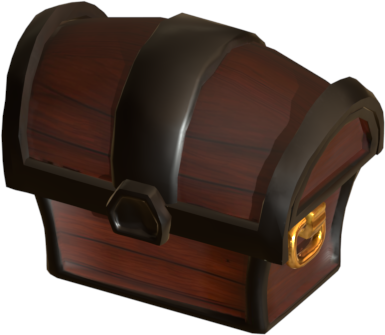} &
    \includegraphics[width=0.115\textwidth,valign=m]{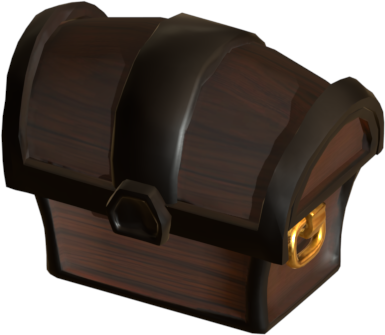} &
    \includegraphics[width=0.115\textwidth,valign=m]{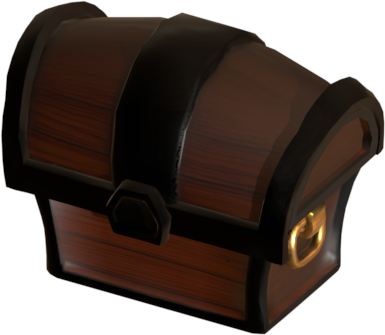}
    \\
    \multicolumn{4}{c}{\textit{\makecell{``A fantasy treasure chest with \receptacle{brown textured wooden planks}, \targetobj{metallic bands}.''}}}\\
    \includegraphics[width=0.115\textwidth,valign=m]{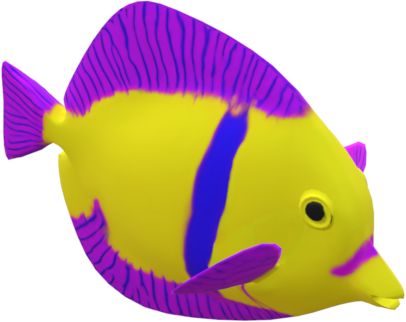} &
    \includegraphics[width=0.115\textwidth,valign=m]{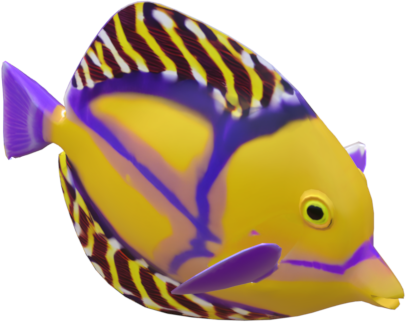} &
    \includegraphics[width=0.115\textwidth,valign=m]{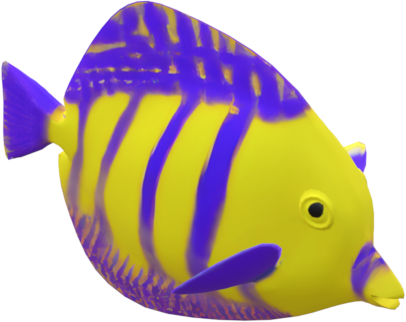} &
    \includegraphics[width=0.115\textwidth,valign=m]{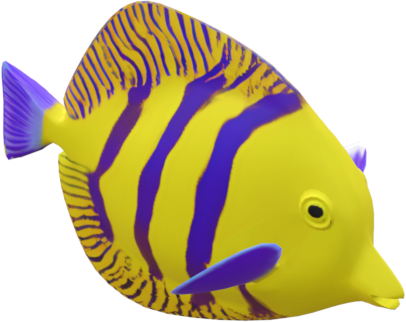}
    \\
    \multicolumn{4}{c}{\textit{\makecell{``A tropical fish with a \receptacle{yellow body}, and \targetobj{purple stripes}.''}}}\\
    \includegraphics[width=0.115\textwidth,valign=m]{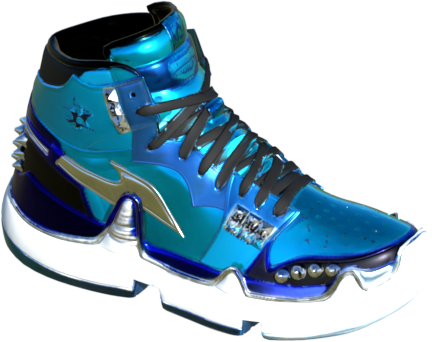} &
    \includegraphics[width=0.115\textwidth,valign=m]{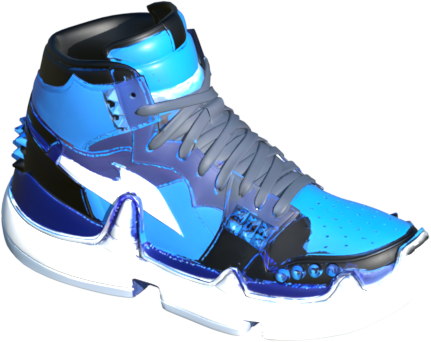} &
    \includegraphics[width=0.115\textwidth,valign=m]{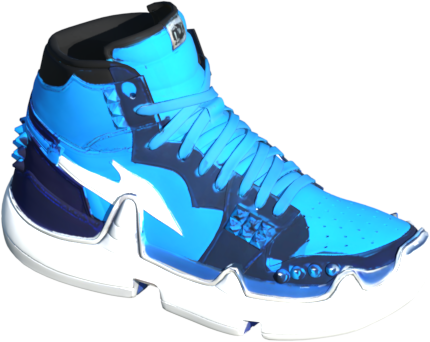} &
    \includegraphics[width=0.115\textwidth,valign=m]{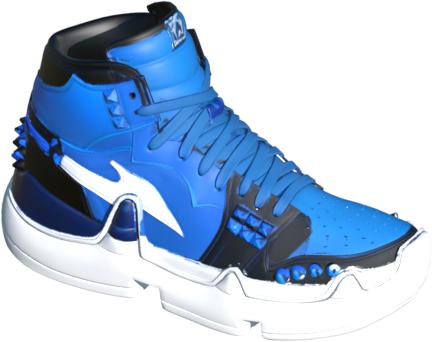}
    \\
    \multicolumn{4}{c}{\textit{\makecell{``A high-top sneaker with \receptacle{materials like leather and fabric}, \targetobj{detailed blue overlays}.''}}}\\
  \end{tabularx}

  \caption{
    \textbf{Qualitative Comparisons for Each Component.}
    Combining residual prediction with $\mathcal{L}_\text{reg}$ effectively preserves the pretrained latent distribution and yields high-quality PBR textures. 
}
  \label{fig:ablation_qualitative}
  \vspace{-1.4\baselineskip}
\end{figure}

\vspace{-\baselineskip}
\paragraph{Evaluation Metrics.} 
Following the previous work~\cite{Fei:2025pacture}, for shaded and albedo images, we report FID~\cite{Heusel:2017fid} and KID~\cite{Bińkowski:2018kid} between generated and reference sets to evaluate fidelity and diversity, with FID computed in the CLIP embedding space~\cite{Radford:2021clip}. 
We additionally measure text–image alignment using CLIP cosine similarity between rendered images and the input prompt. 
For roughness and metallic images, we compute the root mean square error (RMSE) against the corresponding ground-truth images. 
However, these metrics do not reflect the consistency of the generated multi-view material images. 
To this end, we introduce c-PSNR, which computes the PSNR between each pixel $u$ and its corresponding points $\mathcal{C}(u)$. A detailed formulation is provided in the appendix (Sec.~\ref{subsec:cpsnr}). 
Lastly, we report the runtime in seconds required to generate a PBR texture.

\vspace{-0.1\baselineskip}
\subsection{Ablation Studies}
\label{subsec:ablation}
\vspace{-0.25\baselineskip}
In this section, we validate the effectiveness of each component introduced in Sec.~\ref{sec:method}. 
Specifically, we report the following cases discussed in Sec.~\ref{subsec:mcvae}:
\begin{itemize}
    \item \textbf{Frozen VAE~\cite{He:2025materialmvp}}: Directly uses the pretrained VAE, with separate diffusion branches and cross-domain attention.
    \item \textbf{Res. Pred.} + $\mathbf{\mathcal{L}_\text{reg}}$ (\textbf{Ours}): Fine-tunes the encoder using residual prediction with KL regularization (\Ours{}). 
    \item \textbf{Res. Pred.} + $\mathbf{\mathcal{L}_\text{id}}$~\cite{Zhang:2024layerdiffuse}: Replaces KL regularization of \Ours{} with identity loss, as in LayerDiffuse~\cite{Zhang:2024layerdiffuse}. 
    \item \textbf{Direct Pred.} + $\mathbf{\mathcal{L}_\text{reg}}$~\cite{Krishnan:2025orchid}: Replaces residual prediction of \Ours{} with direct prediction, as in Orchid~\cite{Krishnan:2025orchid}. 
\end{itemize}

For each case, the pretrained VAE and diffusion model are fine-tuned for $200$K and $20$K training steps, respectively. 
As Frozen VAE does not require encoder fine-tuning, we allocate more diffusion fine-tuning steps ($45$K) to match the overall training compute.  
Additionally, all cases share the same diffusion architecture and training scheme as \Ours{}, with both CAA and locality regularization enabled. 
To further analyze these components, we additionally consider the components discussed in Sec.~\ref{subsec:mvdit}:
\begin{itemize}
    \item {\textbf{w/o $\mathcal{L}_\text{local}$}}: Locality regularization removed from \matvae{} fine-tuning stage. 
    \item {\textbf{w/o CAA}}: \space CAA module removed from \Ours{}, replaced with dense view attention. 
\end{itemize}

\vspace{-\baselineskip}
\paragraph{Results.}

In Tab.~\ref{tab:ablation}, we present the quantitative results of our ablation studies. 
First, we observe that Frozen VAE (row 1) consistently yields higher FID scores than our \Ours{} (row 2) for both shaded and albedo images, while achieving slightly better or comparable KID scores. 
Moreover, the noticeable gap in the RMSE of roughness and metallic images suggests that the added channels create out-of-distribution latents, degrading generation quality. 
We also note that \Ours{} is about 3$\times$ faster than Frozen VAE, whose higher runtime comes from using separate branches and additional cross-domain attention layers to model inter-domain correlations~\cite{Li:2025idarb, Zhu:2024mcmat}.
Additionally, we observe that \Ours{} outperforms the other encoder fine-tuning schemes (rows 3--4) adopted from previous works~\cite{Krishnan:2025orchid, Zhang:2024layerdiffuse} in FID scores for both shaded and albedo images, while showing slightly higher KID scores. We provide further discussion of these mixed Inception-based KID results, including additional evaluations using CLIP-based KID, in the appendix (Sec.~\ref{subsec:kid_metric}).
Qualitatively, Fig.~\ref{fig:ablation_qualitative} shows that combining residual prediction with $\mathcal{L}_\text{reg}$ enables our \Ours{} to produce PBR textures with superior text-alignment (\eg, the glossy apple surface in row 1) and sharper fine details (\eg, the fish fin in row 3).%

\newcommand{\methodsep}{\cmidrule(lr){1-11}}

\begin{table*}[ht!]
  \centering
  \small
  \caption{\textbf{Quantitative Comparisons of PBR Texture Generation Methods.} Best scores are \textbf{bold}, and the runner-up scores are \underline{underlined}.}
  \vspace{-0.5\baselineskip}
  \label{tab:main_comparison}
  \resizebox{\textwidth}{!}{
  \begin{tabular}{l c ccc ccc c c c}
    \toprule
    \multirow{2}{*}[-0.6ex]{\textbf{Method}} &
    \multirow{2}{*}[-0.6ex]{\textbf{Type}} &
    \multicolumn{3}{c}{\textbf{Shaded}} &
    \multicolumn{3}{c}{\textbf{Albedo}} &
    \textbf{Rough.} &
    \textbf{Metal.} &
    \multirow{2}{*}{\textbf{Time $\downarrow$}} \\
    \cmidrule(lr){3-5}\cmidrule(lr){6-8}\cmidrule(lr){9-9}\cmidrule(lr){10-10}
    & &
    $\mathrm{FID}_{\text{CLIP}}\downarrow$ & $\mathrm{KID}\downarrow$ & $\mathrm{CLIP}\uparrow$ &
    $\mathrm{FID}_{\text{CLIP}}\downarrow$ & $\mathrm{KID}\downarrow$ & $\mathrm{CLIP}\uparrow$ &
    $\mathrm{RMSE}\downarrow$ & $\mathrm{RMSE}\downarrow$ & \\
    \midrule
    MeshGen~\cite{Chen:2025meshgen}          & SC  & 8.637 & 11.105 & 0.282 & 11.322 & 16.193 & 0.282 & \textbf{0.143} & 0.201 & 195s \\
    TexGaussian~\cite{Xiong:2025texgaussian}      & SC  & 6.025 & 3.571 & 0.301 & 12.119 & 9.381 & 0.299 & \underline{0.145} & 0.243 & 73s \\
    \midrule
    Paint-it~\cite{Youwang:2024paintit}         & SDS  & 8.547 & 5.382 & 0.309 & 13.063 & 10.665 & 0.299 & 0.168 & 0.200 & 1260s \\ 
    DreamMat~\cite{Zhang:2024dreammat}         & SDS  & \underline{5.422} & \underline{2.668} & 0.311 & \underline{9.621} & \underline{6.002} & 0.311 & 0.167 & 0.165 & 2400s \\
    FlashTex~\cite{Deng:2024flashtex}         & SDS  & 7.119 & 5.354 & 0.305 & 12.320 & 10.441 & 0.298 & \textbf{0.143} & 0.186 & 285s \\
    \midrule
    MaterialAnything~\cite{Huang:2024materialanything} & MV  & 6.582 & 5.287 & \underline{0.312} & 12.691 & 9.325 & \textbf{0.317} & 0.233 & 0.200 & 500s \\
    MaterialMVP~\cite{He:2025materialmvp}      & MV  & 6.309 & 5.744 & 0.294 & 9.630 & 8.811 & 0.290 & 0.175 & \textbf{0.133} & \underline{35s} \\ 
    {\Ours{}}~\textbf{(Ours)}                      & MV  & \textbf{3.083} & \textbf{1.327} & \textbf{0.318} & \textbf{4.599} & \textbf{1.574} & \underline{0.314} & 0.158 & \underline{0.134} & \textbf{34s} \\
    \bottomrule
  \end{tabular}
  }
  \label{tab:main}
\end{table*}
\vspace{-0.3\baselineskip}

\begin{figure*}[ht!]
  \tiny{
  \centering
  \renewcommand{\arraystretch}{1.0}
  \setlength{\tabcolsep}{0pt}
  \begin{tabularx}{\linewidth}{
  >{\raggedright\arraybackslash}p{0.12\textwidth}
  Y
  >{\raggedright\arraybackslash}p{0.12\textwidth}
  Y
  >{\raggedright\arraybackslash}p{0.12\textwidth}
  Y
  >{\raggedright\arraybackslash}p{0.12\textwidth}
  Y
  }
  \includegraphics[width=0.112\textwidth,valign=m]{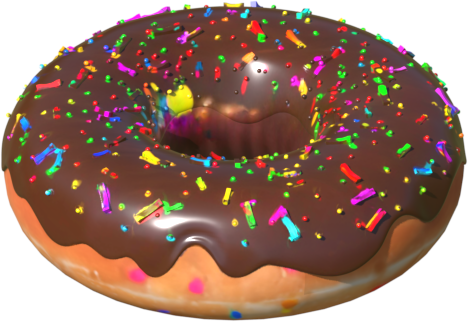} &
  \textit{\makecell{``Realistic donut model\\with a shiny, \receptacle{chocolate}\\\receptacle{-colored frosting},\\\targetobj{multicolored toppings}.''}}&
  \includegraphics[width=0.112\textwidth,valign=m]{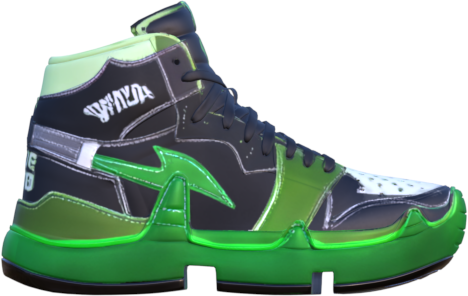} &
  \textit{\makecell{``Detailed sneaker\\model with a \receptacle{black}\\\receptacle{and green} color scheme,\\realistic \targetobj{fabric}\\\targetobj{and leather} textures.''}}&
  \includegraphics[width=0.112\textwidth,valign=m]{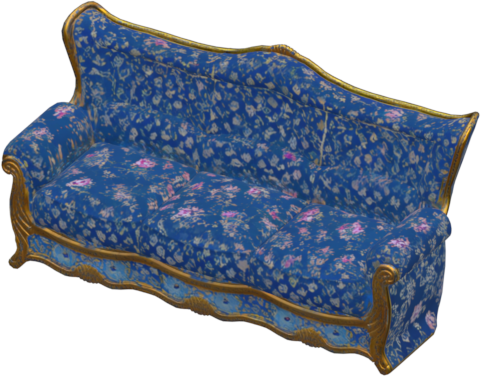} &
  \textit{\makecell{``A \receptacle{blue chintz sofa}\\with a \targetobj{floral pattern,}\\\targetobj{golden edges}.''}}&
  \includegraphics[width=0.112\textwidth,valign=m]{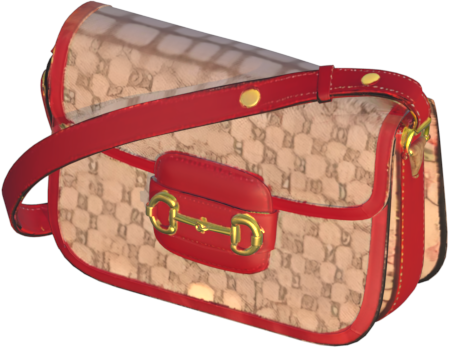} &
  \textit{\makecell{``A classic purse with\\a beige patterned body,\\\receptacle{red leather straps},\\and \targetobj{gold buckles}.''}}\\
  \includegraphics[width=0.112\textwidth,valign=m]{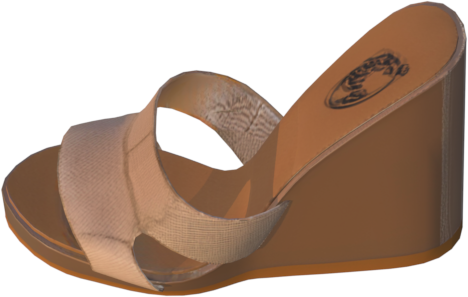} &
  \textit{\makecell{``A fabric-textured sandal\\model with cream color,\\\receptacle{natural leather accents},\\and a \targetobj{distinct branded}\\\targetobj{footbed}.''}}&
  \includegraphics[width=0.112\textwidth,valign=m]{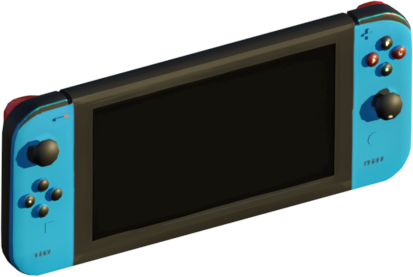} &
  \textit{\makecell{``A \receptacle{realistic handheld}\\\receptacle{gaming console} with\\a \targetobj{black screen, glossy}\\\targetobj{finish}.''}}&
  \includegraphics[width=0.112\textwidth,valign=m]{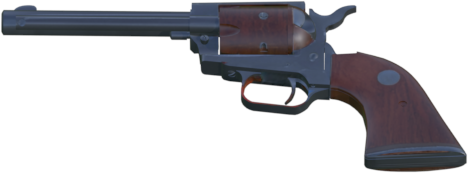} &
  \textit{\makecell{``A detailed revolver with\\a \receptacle{wood-textured grip}, and\\a \targetobj{metallic barrel and}\\\targetobj{slide}.''}}&
  \includegraphics[width=0.112\textwidth,valign=m]{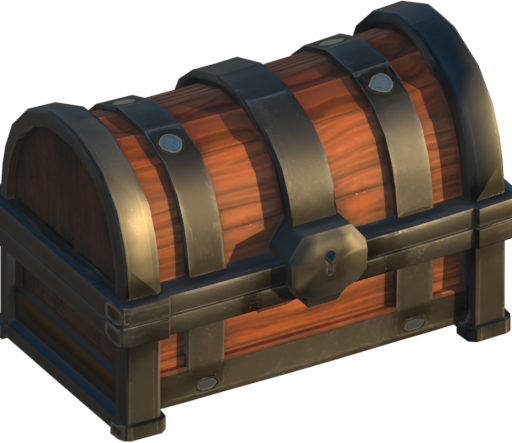} &
  \textit{\makecell{``A fantasy-style chest\\with \receptacle{warm brown wood}\\\receptacle{texture and grey metal}\\\receptacle{frame}, complete with\\ \targetobj{metal bands and rivets}.''}}\\
  \end{tabularx}
  }
  \vspace{-0.5\baselineskip}
  \caption{\textbf{Qualitative Results of \Ours.} 
  \Ours~produces realistic textures with strong text-alignment across diverse material types.}
  \label{fig:main_quali}
  \vspace{-1.5\baselineskip}
\end{figure*}

Next, we ablate the two components introduced in Sec.~\ref{subsec:mvdit}: CAA and locality regularization $\mathcal{L}_\text{local}$. 
When CAA is applied alone without $\mathcal{L}_\text{local}$ (row 5), we observe consistent drops across most metrics, including a noticeable decrease in c-PSNR compared to \Ours{}. 
Without the regularizer, the latent-pixel mapping is not enforced to be spatially local and applying CAA can propagate information across unrelated regions leading to performance drops. 
On the other hand, when CAA is removed and $\mathcal{L}_\text{local}$ is applied alone (row 6), the model does not exhibit noticeable performance drops on most metrics, but still shows inferior c-PSNR, as it must infer correspondences implicitly via dense multi-view attention, resulting in weaker cross-view alignment. 
Our \Ours{}, which combines CAA with $\mathcal{L}_\text{local}$, enforces cross-view feature exchange along geometrically valid correspondences and achieves the highest c-PSNR without performance degradation.

\vspace{-0.25\baselineskip}
\subsection{Comparison to Other Baselines} 
\label{subsec:external_comp} 
\vspace{-0.25\baselineskip}
In this section, we compare \Ours{} to baselines that address PBR texture generation. 
Because prior work differs substantially in training schemes, pretrained models, datasets, and implementation details, a strictly unified experimental setup is impractical. We therefore use each baseline's official checkpoint with its default configuration to generate textured meshes, then render and evaluate all methods under a shared setup.

\vspace{-\baselineskip}
\paragraph{Baselines.}
We compare \Ours{} against the following baseline categories: (i) models trained from scratch to predict PBR textures (\textbf{SC})~\cite{Chen:2025meshgen, Xiong:2025texgaussian}, (ii) SDS-based optimization methods (\textbf{SDS})~\cite{Youwang:2024paintit, Zhang:2024dreammat, Deng:2024flashtex}, and (iii) multi-view diffusion models (\textbf{MV})~\cite{He:2025materialmvp, Huang:2024materialanything}.

\vspace{-\baselineskip}
\paragraph{Results.}
In Tab.~\ref{tab:main}, we present quantitative results of the baselines and \Ours{}. 
We observe that models trained from scratch (SC)~\cite{Chen:2025meshgen, Xiong:2025texgaussian} generally exhibit suboptimal performance compared to \Ours{}, indicating low-fidelity PBR textures caused by suboptimal training under limited high-quality PBR supervision. 
SDS-based methods (SDS)~\cite{Youwang:2024paintit, Zhang:2024dreammat, Deng:2024flashtex} improve over the SC baselines, but the gains are marginal except for DreamMat, which in turn incurs a prohibitive runtime of approximately $\mathbf{40}$ \textbf{minutes}. 
Lastly, \Ours{} effectively leverages the pretrained prior while preserving multi-view consistency, and achieves the \emph{best} performance on shaded and albedo images across all baselines, with the exception of the albedo CLIP score which is marginally behind MaterialAnything~\cite{Huang:2024materialanything}.

We present qualitative results of \Ours{} in Fig.~\ref{fig:main_quali}, which produces high-quality, physically plausible textures across diverse material types. 
We provide comprehensive qualitative comparisons between the baselines and \Ours{} in the appendix (Sec.~\ref{subsec:supp_baseline}) and on the project page.

\vspace{-0.5\baselineskip}
\section{Conclusion}
\label{sec:conclusion}
\vspace{-0.25\baselineskip}
In this work, we presented \Ours{}, a two-stage pipeline that leverages the priors of pretrained image diffusion models for PBR texture generation. 
Directly extending image diffusion models to PBR material images is challenging due to the domain gap introduced by roughness and metallic channels. 
To address this, we introduce effective latent-space adaptation, \matvae{}, which incorporates PBR material images into the pretrained latent distribution, together with locality regularization that enhances multi-view consistency during correspondence-aware attention computation. 
Experimental results show that our \Ours{}, a diffusion model fine-tuned on our material latent space, achieves superior performance in PBR texture generation, with ablations validating the effectiveness of each component.

\subsection*{Acknowledgements}
\vspace{-0.5\baselineskip}
This work was supported by the National Research Foundation of Korea (NRF) (RS-2026-25486000); the Institute of Information \& Communications Technology Planning \& Evaluation (IITP) grants (RS-2024-00399817, RS-2025-25441313, RS-2025-25443318), funded by the Korean government (MSIT); the Industrial Technology Innovation Program (RS-2025-02317326), funded by the Korean government (MOTIE); the National Supercomputing Center (KSC-2025-CRE-0475); and the DRB-KAIST SketchTheFuture Research Center.

{
    \small
    \bibliographystyle{ieeenat_fullname}
    \bibliography{main}

\begin{thebibliography}{64}
\providecommand{\natexlab}[1]{#1}
\providecommand{\url}[1]{\texttt{#1}}
\expandafter\ifx\csname urlstyle\endcsname\relax
  \providecommand{\doi}[1]{doi: #1}\else
  \providecommand{\doi}{doi: \begingroup \urlstyle{rm}\Url}\fi

\bibitem[Aliev et~al.(2025)Aliev, Baranchuk, and Struminsky]{Aliev:2025castex}
Mishan Aliev, Dmitry Baranchuk, and Kirill Struminsky.
\newblock Castex: Cascaded text-to-texture synthesis via explicit texture maps
  and physically-based shading.
\newblock \emph{arXiv preprint arXiv:2504.06856}, 2025.

\bibitem[Bensadoun et~al.(2024)Bensadoun, Kleiman, Azuri, Harosh, Vedaldi,
  Neverova, and Gafni]{bensadoun2024meta}
Raphael Bensadoun, Yanir Kleiman, Idan Azuri, Omri Harosh, Andrea Vedaldi,
  Natalia Neverova, and Oran Gafni.
\newblock Meta {3D} {TextureGen}: Fast and consistent texture generation for
  {3D} objects.
\newblock \emph{arXiv preprint arXiv:2407.02430}, 2024.

\bibitem[Bińkowski et~al.(2018)Bińkowski, Sutherland, Arbel, and
  Gretton]{Bińkowski:2018kid}
Mikołaj Bińkowski, Danica~J. Sutherland, Michael Arbel, and Arthur Gretton.
\newblock Demystifying mmd gans.
\newblock In \emph{ICLR}, 2018.

\bibitem[Byung-Ki et~al.(2025)Byung-Ki, Dai, Hyoseok, Luo, and
  Oh]{Kwon:2025jointdit}
Kwon Byung-Ki, Qi Dai, Lee Hyoseok, Chong Luo, and Tae-Hyun Oh.
\newblock Jointdit: Enhancing rgb-depth joint modeling with diffusion
  transformers.
\newblock In \emph{ICCV}, 2025.

\bibitem[Chen et~al.(2023{\natexlab{a}})Chen, Siddiqui, Lee, Tulyakov, and
  Nießner]{Chen:2023text2tex}
Dave~Zhenyu Chen, Yawar Siddiqui, Hsin-Ying Lee, Sergey Tulyakov, and Matthias
  Nießner.
\newblock Text2tex: Text-driven texture synthesis via diffusion models.
\newblock In \emph{ICCV}, 2023{\natexlab{a}}.

\bibitem[Chen et~al.(2023{\natexlab{b}})Chen, Chen, Jiao, and
  Jia]{Chen:2023fantasia3d}
Rui Chen, Yongwei Chen, Ningxin Jiao, and Kui Jia.
\newblock Fantasia3d: Disentangling geometry and appearance for high-quality
  text-to-3d content creation.
\newblock In \emph{ICCV}, 2023{\natexlab{b}}.

\bibitem[Chen et~al.(2025)Chen, Wang, Sun, Wang, Chen, and
  Liu]{Chen:2025meshgen}
Zilong Chen, Yikai Wang, Wenqiang Sun, Feng Wang, Yiwen Chen, and Huaping Liu.
\newblock Meshgen: Generating pbr textured mesh with render-enhanced
  auto-encoder and generative data augmentation.
\newblock In \emph{CVPR}, 2025.

\bibitem[Deitke et~al.(2023{\natexlab{a}})Deitke, Liu, Wallingford, Ngo,
  Michel, Kusupati, Fan, Laforte, Voleti, Gadre, VanderBilt, Kembhavi,
  Vondrick, Gkioxari, Ehsani, Schmidt, and Farhadi]{Deitke:2023objaversexl}
Matt Deitke, Ruoshi Liu, Matthew Wallingford, Huong Ngo, Oscar Michel, Aditya
  Kusupati, Alan Fan, Christian Laforte, Vikram Voleti, Samir~Yitzhak Gadre,
  Eli VanderBilt, Aniruddha Kembhavi, Carl Vondrick, Georgia Gkioxari, Kiana
  Ehsani, Ludwig Schmidt, and Ali Farhadi.
\newblock Objaverse-xl: A universe of 10m+ 3d objects.
\newblock In \emph{NeurIPS}, 2023{\natexlab{a}}.

\bibitem[Deitke et~al.(2023{\natexlab{b}})Deitke, Schwenk, Salvador, Weihs,
  Michel, VanderBilt, Schmidt, Ehsani, Kembhavi, and
  Farhadi]{deitke2023objaverse}
Matt Deitke, Dustin Schwenk, Jordi Salvador, Luca Weihs, Oscar Michel, Eli
  VanderBilt, Ludwig Schmidt, Kiana Ehsani, Aniruddha Kembhavi, and Ali
  Farhadi.
\newblock Objaverse: A universe of annotated 3d objects.
\newblock In \emph{CVPR}, 2023{\natexlab{b}}.

\bibitem[Deng et~al.(2024)Deng, Omernick, Weiss, Ramanan, Zhu, Zhou, and
  Agrawala]{Deng:2024flashtex}
Kangle Deng, Timothy Omernick, Alexander Weiss, Deva Ramanan, Jun-Yan Zhu,
  Tinghui Zhou, and Maneesh Agrawala.
\newblock Flashtex: Fast relightable mesh texturing with lightcontrolnet.
\newblock In \emph{ECCV}, 2024.

\bibitem[Esser et~al.(2024)Esser, Kulal, Blattmann, Entezari, M{\"u}ller,
  Saini, Levi, Lorenz, Sauer, Boesel, et~al.]{esser2024scaling}
Patrick Esser, Sumith Kulal, Andreas Blattmann, Rahim Entezari, Jonas
  M{\"u}ller, Harry Saini, Yam Levi, Dominik Lorenz, Axel Sauer, Frederic
  Boesel, et~al.
\newblock Scaling rectified flow transformers for high-resolution image
  synthesis.
\newblock In \emph{ICML}, 2024.

\bibitem[Fei et~al.(2025)Fei, Tang, Tian, Shi, and Tan]{Fei:2025pacture}
Fan Fei, Jiajun Tang, Fei-Peng Tian, Boxin Shi, and Ping Tan.
\newblock Pacture: Efficient pbr texture generation on packed views with visual
  autoregressive models.
\newblock \emph{arXiv preprint arXiv:2505.22394}, 2025.

\bibitem[Feng et~al.(2025)Feng, Yang, Yang, Zhang, Yu, Zhao, Liu, Jiang, and
  Guo]{feng2025romantex}
Yifei Feng, Mingxin Yang, Shuhui Yang, Sheng Zhang, Jiaao Yu, Zibo Zhao, Yuhong
  Liu, Jie Jiang, and Chunchao Guo.
\newblock {RomanTex}: Decoupling {3D}-aware rotary positional embedded
  multi-attention network for texture synthesis.
\newblock In \emph{ICCV}, 2025.

\bibitem[Fu et~al.(2024)Fu, Yin, Hu, Wang, Ma, Tan, Shen, Lin, and
  Long]{Fu:2024geowizard}
Xiao Fu, Wei Yin, Mu Hu, Kaixuan Wang, Yuexin Ma, Ping Tan, Shaojie Shen, Dahua
  Lin, and Xiaoxiao Long.
\newblock Geowizard: Unleashing the diffusion priors for 3d geometry estimation
  from a single image.
\newblock In \emph{ECCV}, 2024.

\bibitem[Gao et~al.(2024)Gao, Holynski, Henzler, Brussee, Martin-Brualla,
  Srinivasan, Barron, and Poole]{Gao:2024cat3d}
Ruiqi Gao, Aleksander Holynski, Philipp Henzler, Arthur Brussee, Ricardo
  Martin-Brualla, Pratul Srinivasan, Jonathan~T. Barron, and Ben Poole.
\newblock Cat3d: Create anything in 3d with multi-view diffusion models.
\newblock In \emph{NeurIPS}, 2024.

\bibitem[Goodfellow et~al.(2014)Goodfellow, Pouget-Abadie, Mirza, Xu,
  Warde-Farley, Ozair, Courville, and Bengio]{Goodfellow:2014gan}
Ian~J. Goodfellow, Jean Pouget-Abadie, Mehdi Mirza, Bing Xu, David
  Warde-Farley, Sherjil Ozair, Aaron Courville, and Yoshua Bengio.
\newblock Generative adversarial networks.
\newblock In \emph{NeurIPS}, 2014.

\bibitem[Hassan et~al.(2023)Hassan, Guo, Wang, Black, Fidler, and
  Peng]{Hassan:2023synthesizing}
Mohamed Hassan, Yunrong Guo, Tingwu Wang, Michael Black, Sanja Fidler, and
  Xue~Bin Peng.
\newblock Synthesizing physical character-scene interactions.
\newblock In \emph{SIGGRAPH}, 2023.

\bibitem[He et~al.(2025)He, Yang, Yang, Tang, Wang, Zhang, Chen, Liu, Jiang,
  Guo, and Luo]{He:2025materialmvp}
Zebin He, Mingxin Yang, Shuhui Yang, Yixuan Tang, Tao Wang, Kaihao Zhang,
  Guanying Chen, Yuhong Liu, Jie Jiang, Chunchao Guo, and Wenhan Luo.
\newblock Materialmvp: Illumination-invariant material generation via
  multi-view pbr diffusion.
\newblock In \emph{ICCV}, 2025.

\bibitem[Heusel et~al.(2017)Heusel, Ramsauer, Unterthiner, Nessler, and
  Hochreiter]{Heusel:2017fid}
Martin Heusel, Hubert Ramsauer, Thomas Unterthiner, Bernhard Nessler, and Sepp
  Hochreiter.
\newblock Gans trained by a two time-scale update rule converge to a local nash
  equilibrium.
\newblock In \emph{NeurIPS}, 2017.

\bibitem[Ho and Salimans(2022)]{ho2022classifier}
Jonathan Ho and Tim Salimans.
\newblock Classifier-free diffusion guidance.
\newblock \emph{arXiv preprint arXiv:2207.12598}, 2022.

\bibitem[Hu et~al.(2022)Hu, Shen, Wallis, Allen-Zhu, Li, Wang, Wang, Chen,
  et~al.]{hu2022lora}
Edward~J Hu, Yelong Shen, Phillip Wallis, Zeyuan Allen-Zhu, Yuanzhi Li, Shean
  Wang, Lu Wang, Weizhu Chen, et~al.
\newblock {LoRA}: Low-rank adaptation of large language models.
\newblock In \emph{ICLR}, 2022.

\bibitem[Huang et~al.(2025{\natexlab{a}})Huang, Wang, Liu, and
  Wang]{Huang:2024materialanything}
Xin Huang, Tengfei Wang, Ziwei Liu, and Qing Wang.
\newblock Material anything: Generating materials for any 3d object via
  diffusion.
\newblock In \emph{CVPR}, 2025{\natexlab{a}}.

\bibitem[Huang et~al.(2025{\natexlab{b}})Huang, Guo, Wang, Yi, Ma, Cao, and
  Sheng]{Huang:2025mvadapter}
Zehuan Huang, Yuan-Chen Guo, Haoran Wang, Ran Yi, Lizhuang Ma, Yan-Pei Cao, and
  Lu Sheng.
\newblock Mv-adapter: Multi-view consistent image generation made easy.
\newblock In \emph{ICCV}, 2025{\natexlab{b}}.

\bibitem[Höllein et~al.(2023)Höllein, Cao, Owens, Johnson, and
  Nießner]{Höllein:2023text2room}
Lukas Höllein, Ang Cao, Andrew Owens, Justin Johnson, and Matthias Nießner.
\newblock Text2room: Extracting textured 3d meshes from 2d text-to-image
  models.
\newblock In \emph{ICCV}, 2023.

\bibitem[Jayasumana et~al.(2024)Jayasumana, Ramalingam, Veit, Glasner,
  Chakrabarti, and Kumar]{jayasumana2024rethinking}
Sadeep Jayasumana, Srikumar Ramalingam, Andreas Veit, Daniel Glasner, Ayan
  Chakrabarti, and Sanjiv Kumar.
\newblock Rethinking fid: Towards a better evaluation metric for image
  generation.
\newblock In \emph{CVPR}, 2024.

\bibitem[Jun and Nichol(2023)]{Jun:2023shap-e}
Heewoo Jun and Alex Nichol.
\newblock Shap-e: Generating conditional 3d implicit functions.
\newblock \emph{arXiv preprint arXiv:2305.02463}, 2023.

\bibitem[Kant et~al.(2024)Kant, Wu, Vasilkovsky, Qian, Ren, Guler, Ghanem,
  Tulyakov, Gilitschenski, and Siarohin]{Kant:2024spad}
Yash Kant, Ziyi Wu, Michael Vasilkovsky, Guocheng Qian, Jian Ren, Riza~Alp
  Guler, Bernard Ghanem, Sergey Tulyakov, Igor Gilitschenski, and Aliaksandr
  Siarohin.
\newblock {SPAD}: Spatially aware multiview diffusers.
\newblock In \emph{CVPR}, 2024.

\bibitem[Kim et~al.(2024)Kim, Koo, Yeo, and Sung]{Kim:2024synctweedies}
Jaihoon Kim, Juil Koo, Kyeongmin Yeo, and Minhyuk Sung.
\newblock Synctweedies: A general generative framework based on synchronized
  diffusions.
\newblock In \emph{NeurIPS}, 2024.

\bibitem[Krishnan et~al.(2025)Krishnan, Yan, Casser, and
  Kundu]{Krishnan:2025orchid}
Akshay Krishnan, Xinchen Yan, Vincent Casser, and Abhijit Kundu.
\newblock Orchid: Image latent diffusion for joint appearance and geometry
  generation.
\newblock In \emph{ICCV}, 2025.

\bibitem[Kynk{\"a}{\"a}nniemi et~al.(2023)Kynk{\"a}{\"a}nniemi, Karras,
  Aittala, Aila, and Lehtinen]{kynkaanniemirole}
Tuomas Kynk{\"a}{\"a}nniemi, Tero Karras, Miika Aittala, Timo Aila, and Jaakko
  Lehtinen.
\newblock The role of imagenet classes in fr{\'e}chet inception distance.
\newblock In \emph{ICLR}, 2023.

\bibitem[Li et~al.(2024{\natexlab{a}})Li, Shi, Zhang, Wu, Liao, Wang, Lee, and
  Zhou]{Li:2024dreamscene}
Haoran Li, Haolin Shi, Wenli Zhang, Wenjun Wu, Yong Liao, Lin Wang, Lik-Hang
  Lee, and Peng~Yuan Zhou.
\newblock Dreamscene: 3d gaussian-based text-to-3d scene generation via
  formation pattern sampling.
\newblock In \emph{ECCV}, 2024{\natexlab{a}}.

\bibitem[Li et~al.(2024{\natexlab{b}})Li, Liu, Long, Zhang, Lin, Li, Qi, Zhang,
  Luo, Tan, Wang, Liu, and Guo]{Li:2024era3d}
Peng Li, Yuan Liu, Xiaoxiao Long, Feihu Zhang, Cheng Lin, Mengfei Li, Xingqun
  Qi, Shanghang Zhang, Wenhan Luo, Ping Tan, Wenping Wang, Qifeng Liu, and Yike
  Guo.
\newblock Era3d: High-resolution multiview diffusion using efficient row-wise
  attention.
\newblock In \emph{NeurIPS}, 2024{\natexlab{b}}.

\bibitem[Li et~al.(2025)Li, Wu, Tan, Zhang, Wang, and Lin]{Li:2025idarb}
Zhibing Li, Tong Wu, Jing Tan, Mengchen Zhang, Jiaqi Wang, and Dahua Lin.
\newblock Idarb: Intrinsic decomposition for arbitrary number of input views
  and illuminations.
\newblock In \emph{ICLR}, 2025.

\bibitem[Lin et~al.(2023)Lin, Gao, Tang, Takikawa, Zeng, Huang, Kreis, Fidler,
  Liu, and Lin]{Lin:2023magic3d}
Chen-Hsuan Lin, Jun Gao, Luming Tang, Towaki Takikawa, Xiaohui Zeng, Xun Huang,
  Karsten Kreis, Sanja Fidler, Ming-Yu Liu, and Tsung-Yi Lin.
\newblock Magic3d: High-resolution text-to-3d content creation.
\newblock In \emph{CVPR}, 2023.

\bibitem[Lipman et~al.(2023)Lipman, Chen, Ben-Hamu, Nickel, and
  Le]{Lipman:2023flowmatching}
Yaron Lipman, Ricky T.~Q. Chen, Heli Ben-Hamu, Maximilian Nickel, and Matt Le.
\newblock Flow matching for generative modeling.
\newblock In \emph{ICLR}, 2023.

\bibitem[Long et~al.(2024)Long, Guo, Lin, Liu, Dou, Liu, Ma, Zhang, Habermann,
  Theobalt, and Wang]{Long:2023wonder3d}
Xiaoxiao Long, Yuan-Chen Guo, Cheng Lin, Yuan Liu, Zhiyang Dou, Lingjie Liu,
  Yuexin Ma, Song-Hai Zhang, Marc Habermann, Christian Theobalt, and Wenping
  Wang.
\newblock Wonder3d: Single image to 3d using cross-domain diffusion.
\newblock In \emph{CVPR}, 2024.

\bibitem[Luo et~al.(2023)Luo, Rockwell, Lee, and Johnson]{Luo:2023cap3d}
Tiange Luo, Chris Rockwell, Honglak Lee, and Justin Johnson.
\newblock Scalable 3d captioning with pretrained models.
\newblock In \emph{NeurIPS}, 2023.

\bibitem[Petrovich et~al.(2023)Petrovich, Black, and Varol]{Petrovich:2023tmr}
Mathis Petrovich, Michael~J. Black, and Gül Varol.
\newblock Tmr: Text-to-motion retrieval using contrastive 3d human motion
  synthesis.
\newblock In \emph{ICCV}, 2023.

\bibitem[Poole et~al.(2023)Poole, Jain, Barron, and
  Mildenhall]{Poole:2022dreamfusion}
Ben Poole, Ajay Jain, Jonathan~T. Barron, and Ben Mildenhall.
\newblock Dreamfusion: Text-to-3d using 2d diffusion.
\newblock In \emph{ICLR}, 2023.

\bibitem[Radford et~al.(2021)Radford, Kim, Hallacy, Ramesh, Goh, Agarwal,
  Sastry, Askell, Mishkin, Clark, Krueger, and Sutskever]{Radford:2021clip}
Alec Radford, Jong~Wook Kim, Chris Hallacy, Aditya Ramesh, Gabriel Goh,
  Sandhini Agarwal, Girish Sastry, Amanda Askell, Pamela Mishkin, Jack Clark,
  Gretchen Krueger, and Ilya Sutskever.
\newblock Learning transferable visual models from natural language
  supervision.
\newblock In \emph{ICML}, 2021.

\bibitem[Richardson et~al.(2023)Richardson, Metzer, Alaluf, Giryes, and
  Cohen-Or]{Richardson:2023texture}
Elad Richardson, Gal Metzer, Yuval Alaluf, Raja Giryes, and Daniel Cohen-Or.
\newblock Texture: Text-guided texturing of 3d shapes.
\newblock In \emph{SIGGRAPH}, 2023.

\bibitem[Schuhmann et~al.(2022)Schuhmann, Beaumont, Vencu, Gordon, Wightman,
  Cherti, Coombes, Katta, Mullis, Wortsman, Schramowski, Kundurthy, Crowson,
  Schmidt, Kaczmarczyk, and Jitsev]{Schuhmann:2022laion5b}
Christoph Schuhmann, Romain Beaumont, Richard Vencu, Cade Gordon, Ross
  Wightman, Mehdi Cherti, Theo Coombes, Aarush Katta, Clayton Mullis, Mitchell
  Wortsman, Patrick Schramowski, Srivatsa Kundurthy, Katherine Crowson, Ludwig
  Schmidt, Robert Kaczmarczyk, and Jenia Jitsev.
\newblock Laion-5b: An open large-scale dataset for training next generation
  image-text models.
\newblock In \emph{NeurIPS}, 2022.

\bibitem[Shi et~al.(2024)Shi, Wang, Ye, Long, Li, and Yang]{Shi:2024mvdream}
Yichun Shi, Peng Wang, Jianglong Ye, Mai Long, Kejie Li, and Xiao Yang.
\newblock Mvdream: Multi-view diffusion for 3d generation.
\newblock In \emph{ICLR}, 2024.

\bibitem[Siddiqui et~al.(2024)Siddiqui, Monnier, Kokkinos, Kariya, Kleiman,
  Garreau, Gafni, Neverova, Vedaldi, Shapovalov, and
  Novotny]{Siddiqui:2024assetgen}
Yawar Siddiqui, Tom Monnier, Filippos Kokkinos, Mahendra Kariya, Yanir Kleiman,
  Emilien Garreau, Oran Gafni, Natalia Neverova, Andrea Vedaldi, Roman
  Shapovalov, and David Novotny.
\newblock Meta 3d assetgen: Text-to-mesh generation with high-quality geometry,
  texture, and pbr materials.
\newblock In \emph{NeurIPS}, 2024.

\bibitem[Sun et~al.(2025)Sun, Wu, Zhang, Zang, Dong, Xiong, Lin, and
  Wang]{Sun:2025bootstrap3d}
Zeyi Sun, Tong Wu, Pan Zhang, Yuhang Zang, Xiaoyi Dong, Yuanjun Xiong, Dahua
  Lin, and Jiaqi Wang.
\newblock Bootstrap3d: Improving multi-view diffusion model with synthetic
  data.
\newblock In \emph{ICCV}, 2025.

\bibitem[Tang et~al.(2023)Tang, Zhang, Chen, Wang, and
  Furukawa]{Tang:2023mvdiffusion}
Shitao Tang, Fuyang Zhang, Jiacheng Chen, Peng Wang, and Yasutaka Furukawa.
\newblock Mvdiffusion: Enabling holistic multi-view image generation with
  correspondence-aware diffusion.
\newblock In \emph{NeurIPS}, 2023.

\bibitem[Tevet et~al.(2023)Tevet, Raab, Gordon, Shafir, Cohen-Or, and
  Bermano]{Tevet:2023humanmotiondiffusionmodel}
Guy Tevet, Sigal Raab, Brian Gordon, Yonatan Shafir, Daniel Cohen-Or, and
  Amit~H. Bermano.
\newblock Human motion diffusion model.
\newblock In \emph{ICLR}, 2023.

\bibitem[Wang and Shi(2023)]{Wang:2023imagedream}
Peng Wang and Yichun Shi.
\newblock Imagedream: Image-prompt multi-view diffusion for 3d generation.
\newblock \emph{arXiv preprint arXiv:2312.02201}, 2023.

\bibitem[Wang et~al.(2024)Wang, Xu, Ma, Wang, and Dai]{Wang:2024boosting3d}
Yitong Wang, Xudong Xu, Li Ma, Haoran Wang, and Bo Dai.
\newblock Boosting 3d object generation through pbr materials.
\newblock In \emph{SIGGRAPH Asia}, 2024.

\bibitem[Wang et~al.(2023)Wang, Lu, Wang, Bao, Li, Su, and
  Zhu]{Wang:2023prolificdreamer}
Zhengyi Wang, Cheng Lu, Yikai Wang, Fan Bao, Chongxuan Li, Hang Su, and Jun
  Zhu.
\newblock Prolificdreamer: High-fidelity and diverse text-to-3d generation with
  variational score distillation.
\newblock In \emph{NeurIPS}, 2023.

\bibitem[Wei et~al.(2025)Wei, Zhang, Yang, Wang, Guo, Zhao, and
  Luximon]{Wei:2025pbr3dgen}
Xiaokang Wei, Bowen Zhang, Xianghui Yang, Yuxuan Wang, Chunchao Guo, Xi Zhao,
  and Yan Luximon.
\newblock Pbr3dgen: A vlm-guided mesh generation with high-quality pbr texture.
\newblock \emph{arXiv preprint arXiv:2503.11368}, 2025.

\bibitem[Xiong et~al.(2025)Xiong, Liu, Hu, Wu, Wu, Liu, Zhao, Ding, and
  Lian]{Xiong:2025texgaussian}
Bojun Xiong, Jialun Liu, Jiakui Hu, Chenming Wu, Jinbo Wu, Xing Liu, Chen Zhao,
  Errui Ding, and Zhouhui Lian.
\newblock Texgaussian: Generating high-quality pbr material via octree-based 3d
  gaussian splatting.
\newblock In \emph{CVPR}, 2025.

\bibitem[Yeo et~al.(2025)Yeo, Kim, and Sung]{Yeo:2025stochsync}
Kyeongmin Yeo, Jaihoon Kim, and Minhyuk Sung.
\newblock Stochsync: Stochastic diffusion synchronization for image generation
  in arbitrary spaces.
\newblock In \emph{ICLR}, 2025.

\bibitem[Yi et~al.(2024)Yi, Fang, Wang, Wu, Xie, Zhang, Liu, Tian, and
  Wang]{Yi:2024gaussiandreamer}
Taoran Yi, Jiemin Fang, Junjie Wang, Guanjun Wu, Lingxi Xie, Xiaopeng Zhang,
  Wenyu Liu, Qi Tian, and Xinggang Wang.
\newblock Gaussiandreamer: Fast generation from text to 3d gaussians by
  bridging 2d and 3d diffusion models.
\newblock In \emph{CVPR}, 2024.

\bibitem[Youwang et~al.(2024)Youwang, Oh, and Pons-Moll]{Youwang:2024paintit}
Kim Youwang, Tae-Hyun Oh, and Gerard Pons-Moll.
\newblock Paint-it: Text-to-texture synthesis via deep convolutional texture
  map optimization and physically-based rendering.
\newblock In \emph{CVPR}, 2024.

\bibitem[Yu et~al.(2025)Yu, Gu, Hu, Li, and Dong]{Yu:2025unicon}
Fanghua Yu, Jinjin Gu, Jinfan Hu, Zheyuan Li, and Chao Dong.
\newblock Unicon: Unidirectional information flow for effective control of
  large-scale diffusion models.
\newblock In \emph{ICLR}, 2025.

\bibitem[Zeng et~al.(2024)Zeng, Deschaintre, Georgiev, Hold-Geoffroy, Hu, Luan,
  Yan, and Hašan]{Zeng:2024rgbx}
Zheng Zeng, Valentin Deschaintre, Iliyan Georgiev, Yannick Hold-Geoffroy, Yiwei
  Hu, Fujun Luan, Ling-Qi Yan, and Miloš Hašan.
\newblock {RGB$\leftrightarrow$X}: Image decomposition and synthesis using
  material- and lighting-aware diffusion models.
\newblock In \emph{SIGGRAPH}, 2024.

\bibitem[Zhang and Agrawala(2024)]{Zhang:2024layerdiffuse}
Lvmin Zhang and Maneesh Agrawala.
\newblock Transparent image layer diffusion using latent transparency.
\newblock In \emph{SIGGRAPH}, 2024.

\bibitem[Zhang et~al.(2023)Zhang, Rao, and Agrawala]{zhang2023adding}
Lvmin Zhang, Anyi Rao, and Maneesh Agrawala.
\newblock Adding conditional control to text-to-image diffusion models.
\newblock In \emph{ICCV}, 2023.

\bibitem[Zhang et~al.(2024{\natexlab{a}})Zhang, Wang, Siarohin, Zhuang, Xu,
  Yang, Lin, Zhou, Tulyakov, and Lee]{Zhang:2024scenewiz3d}
Qihang Zhang, Chaoyang Wang, Aliaksandr Siarohin, Peiye Zhuang, Yinghao Xu,
  Ceyuan Yang, Dahua Lin, Bolei Zhou, Sergey Tulyakov, and Hsin-Ying Lee.
\newblock Scenewiz3d: Towards text-guided 3d scene composition.
\newblock In \emph{CVPR}, 2024{\natexlab{a}}.

\bibitem[Zhang et~al.(2024{\natexlab{b}})Zhang, Liu, Xie, Yang, Liu, Yang,
  Zhang, Kou, Lin, Wang, and Jin]{Zhang:2024dreammat}
Yuqing Zhang, Yuan Liu, Zhiyu Xie, Lei Yang, Zhongyuan Liu, Mengzhou Yang,
  Runze Zhang, Qilong Kou, Cheng Lin, Wenping Wang, and Xiaogang Jin.
\newblock Dreammat: High-quality pbr material generation with geometry- and
  light-aware diffusion models.
\newblock In \emph{SIGGRAPH}, 2024{\natexlab{b}}.

\bibitem[Zhang et~al.(2025)Zhang, Zhang, Ma, and Cao]{zhang2025texverse}
Yibo Zhang, Li Zhang, Rui Ma, and Nan Cao.
\newblock Texverse: A universe of 3d objects with high-resolution textures.
\newblock \emph{arXiv preprint arXiv:2508.10868}, 2025.

\bibitem[Zhu et~al.(2025)Zhu, Ye, Zhang, Hu, Yin, Li, Chen, Qian, Wang, Liao,
  and Yu]{Zhu:2025muma}
Lingting Zhu, Jingrui Ye, Runze Zhang, Zeyu Hu, Yingda Yin, Lanjiong Li, Jinnan
  Chen, Shengju Qian, Xin Wang, Qingmin Liao, and Lequan Yu.
\newblock Muma: 3d pbr texturing via multi-channel multi-view generation and
  agentic post-processing.
\newblock \emph{arXiv preprint arXiv:2503.18461}, 2025.

\bibitem[Zhu et~al.(2024)Zhu, Qiu, Gu, Zhao, Xu, He, Li, Han, Yao, Cao, Zhu,
  Yuan, Dong, and Zhu]{Zhu:2024mcmat}
Shenhao Zhu, Lingteng Qiu, Xiaodong Gu, Zhengyi Zhao, Chao Xu, Yuxiao He, Zhe
  Li, Xiaoguang Han, Yao Yao, Xun Cao, Siyu Zhu, Weihao Yuan, Zilong Dong, and
  Hao Zhu.
\newblock Mcmat: Multiview-consistent and physically accurate pbr material
  generation.
\newblock \emph{arXiv preprint arXiv:2412.14148}, 2024.

\end{thebibliography}
}

\clearpage
\appendix
\newcommand{\refofpaper}[1]{\unskip}
\newcommand{\refinpaper}[1]{\unskip}

\section*{Appendix}
In this appendix, we present technical discussions (Sec.~\ref{sec:technical_details}), implementation details (Sec.~\ref{sec:impl_details}), and qualitative results (Sec.~\ref{sec:quali_res}). 

\section{Technical Discussions}
\label{sec:technical_details}
In this section, we detail our correspondence-PSNR metric (Sec.~\ref{subsec:cpsnr}), discuss the KID evaluation metric (Sec.~\ref{subsec:kid_metric}), and provide further analyses of VAE prediction types (Sec.~\ref{subsec:predict_type}), locality regularization (Sec.~\ref{subsec:locality_local}), and latent distribution mismatch (Sec.~\ref{subsec:latent_mismatch}).

\subsection{Correspondence-PSNR}
\label{subsec:cpsnr}
To quantify the multi-view consistency of the generated PBR material images, we introduced a correspondence-PSNR (c-PSNR) metric in Sec.~\ref{sec:experiments}. 

Specifically, given a set of $N$ generated view images $\{\mathbf{x}_i\}_{i=1}^N$ and a 3D mesh, for each pixel $u \in \Omega$ in the $i$-th view, the set of corresponding pixels in the $j$-th view is denoted as $\mathcal{C}_{i \rightarrow j}(u)$. We then compute the MSE (mean squared error) as follows:
\begin{equation}
\label{eq:cpsnr}
\text{MSE} = \frac{1}{CM} \sum_{i=1}^N \sum_{j \ne i} \sum_{\substack{u \in \Omega, \\ v \in \mathcal{C}_{i \rightarrow j}(u)}} \|\mathbf{x}_i(u) - \mathbf{x}_j(v)\|^2,
\end{equation}
where $C=5$ is the number of channels of $\mathbf{x}$ and $M = \sum_{i=1}^N \sum_{j \ne i} \sum_{u \in \Omega} |\mathcal{C}_{i \rightarrow j}(u)|$ is the total number of correspondence pairs. 
Accordingly, we define the correspondence-PSNR as
\[
\text{c-PSNR} = 10 \log_{10}(1/\text{MSE}),
\]
which measures the discrepancy across all geometric correspondences between views. 
As shown in Tab.~\ref{tab:ablation}, \Ours{} achieves higher c-PSNR than methods without CAA, indicating more consistent PBR material images across views.

\begin{figure}[t]
  \centering
  \includegraphics[width=0.9\linewidth]{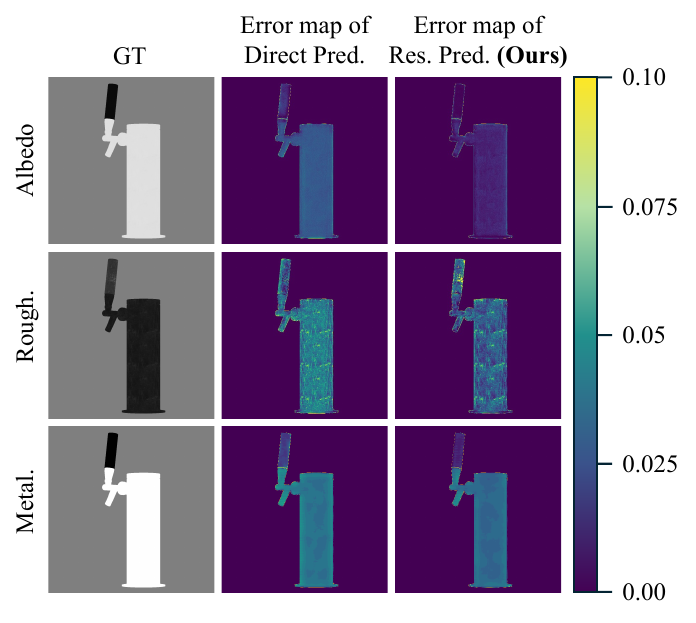}
  \vspace{-0.5\baselineskip}
  \caption{
    \textbf{Direct vs. Residual Prediction.}
    We show the albedo, roughness, and metallic reconstruction errors at an early training stage (10k iterations) for direct and residual prediction. Residual prediction preserves the pretrained latent representation, achieving superior albedo reconstruction quality (darker regions indicate lower errors). 
  }
  \label{fig:supp_residual_errors}
  \vspace{0.0\baselineskip}
\end{figure}

\begin{table}[b]
    \centering
    \small
    \setlength{\tabcolsep}{6pt}
    \renewcommand{\arraystretch}{1.12}
    \begin{tabular}{lcc}
        \toprule
        \multirow{2}{*}{\textbf{Methods}} & \textbf{Shaded} & \textbf{Albedo} \\
        \cmidrule(lr){2-2} \cmidrule(lr){3-3}
         & $\mathrm{KID}_{\text{CLIP}}\downarrow$ & $\mathrm{KID}_{\text{CLIP}}\downarrow$ \\
        \midrule
        Frozen VAE~\cite{He:2025materialmvp} & 2.123 & 3.547 \\
        \textbf{Res. Pred. + $\mathcal{L}_{\text{reg}}$ (Ours)} & \underline{1.496} & \textbf{2.855} \\
        Res. Pred. + $\mathcal{L}_{\text{id}}$~\cite{Zhang:2024layerdiffuse} & 1.977 & 3.710 \\
        Direct Pred. + $\mathcal{L}_{\text{reg}}$~\cite{Krishnan:2025orchid} & 1.917 & 3.254 \\
        w/o $\mathcal{L}_{\text{local}}$ & 2.660 & 8.820 \\
        w/o CAA & \textbf{1.424} & \underline{3.154} \\
        \bottomrule
    \end{tabular}
    \caption{\textbf{Evaluation on CLIP-based KID.} We additionally measure KID in the CLIP feature space. Our proposed VAE fine-tuning scheme yields superior performance, with both our full model and the variant without CAA consistently outperforming the other baselines.}
    \label{tab:kid_clip}
    \vspace{-0.5\baselineskip}
\end{table}

\subsection{Discussion on KID Evaluation}
\label{subsec:kid_metric}
In Tab.~\ref{tab:ablation}, our full model achieves the best $\mathrm{FID}_{\mathrm{CLIP}}$ under the evaluation protocol of~\cite{Fei:2025pacture}, while the results on Inception-based KID are more mixed. We hypothesize that Inception features may not faithfully reflect perceptual similarity for rendered material images, as prior works have shown that they are strongly tied to ImageNet semantics and can diverge from human perception~\cite{kynkaanniemirole, jayasumana2024rethinking}.

To further examine this effect, we additionally report CLIP-based KID in Tab.~\ref{tab:kid_clip}. Compared with the Inception-based KID results in Tab.~\ref{tab:ablation}, CLIP-based KID is more consistent with $\mathrm{FID}_{\mathrm{CLIP}}$ and with recent work that adopts CLIP-based Fréchet or kernel distances for rendered images~\cite{He:2025materialmvp,feng2025romantex}. Under this metric, \Ours{} consistently outperforms the baselines, including the direct adaptations of prior VAE fine-tuning schemes.

\subsection{Additional Analysis of Direct and Residual Prediction}
\label{subsec:predict_type}
As discussed in Sec.~\ref{subsec:mcvae}, residual prediction can be more effective than direct prediction for VAE fine-tuning.
In this section, we present an empirical comparison of residual and direct prediction for VAE during the early stage of fine-tuning, highlighting their optimization behavior and stability.

We present a qualitative comparison in Fig.~\ref{fig:supp_residual_errors}. The left column shows the ground-truth albedo, roughness, and metallic maps. 
The middle and right columns show reconstruction error maps, computed between the ground-truth images and the outputs of VAEs fine-tuned with direct prediction and residual prediction (Ours), respectively, where brighter regions indicate higher errors.
All results are evaluated at the same early fine-tuning step (after 10k iterations) for a fair comparison. 

Note that the reconstruction error of the albedo map under residual prediction is smaller than that under direct prediction. 
This is because residual prediction initializes the final layer weights of the offset encoder to zero, producing zero residuals at the start of training and thereby preserving the original latent representation.
In contrast, direct prediction is not constrained to remain aligned with the pretrained latent space and, even after several update steps, already exhibits noticeable degradation in albedo reconstruction quality.
While the errors for roughness and metallic maps remain comparable between the two schemes, we empirically observe that the improved stability of residual prediction on albedo images translates into better downstream generation performance, as reported in Tab.~\ref{tab:ablation}.

\begin{figure}[t]
  \centering
  \includegraphics[width=1.0\linewidth]{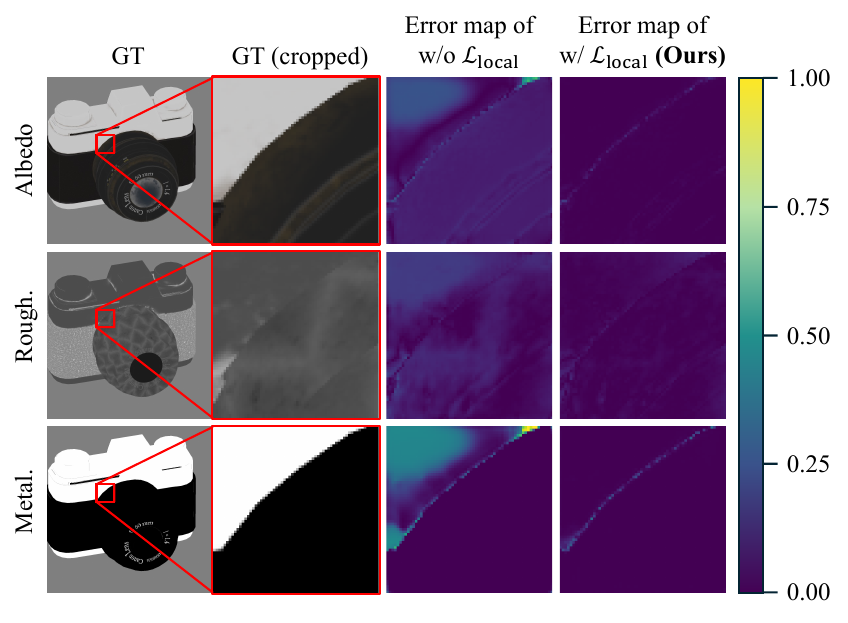}
  \caption{
      \textbf{Visualization of Reconstruction Error Maps on Cropped Patches.}
      We show albedo, roughness, and metallic reconstruction error maps for \textsc{MatVAE} trained with and without $\mathcal{L}_{\text{local}}$. 
      Applying $\mathcal{L}_{\text{local}}$ significantly reduces patch reconstruction errors, indicating improved latent--image spatial alignment.
  }
  \label{fig:supp_crop_errors}
  \vspace{-1.0\baselineskip}
\end{figure}

\subsection{Additional Analysis of Locality Regularization}
\label{subsec:locality_local}
As discussed in Sec.~\ref{subsec:mvdit}, applying locality regularization $\mathcal{L}_{\text{local}}$ improves latent-image spatial alignment and leads to superior performance when combined with CAA. 
In Fig.~\ref{fig:supp_crop_errors}, we visualize the reconstruction error of cropped patches using a VAE fine-tuned with and without $\mathcal{L}_{\text{local}}$ to further analyze the effect of the regularizer. 
Specifically, we show cropped patches from the albedo, roughness, and metallic images, together with their reconstruction error maps, $\lVert \mathbf{x} - \mathcal{D}(\mathcal{E}(\mathbf{x})) \rVert^{2}$, computed at an $8 \times 8$ latent resolution for VAEs fine-tuned without and with $\mathcal{L}_{\text{local}}$, where brighter regions indicate higher errors.

Note that applying $\mathcal{L}_{\text{local}}$ significantly reduces the reconstruction error, indicating that the learned latent representation achieves improved latent--image spatial alignment. 
In contrast, removing $\mathcal{L}_{\text{local}}$ disrupts the spatial alignment, indicating that the fine-tuned encoder entangles information across spatially distant latent tokens. 
Consequently, the latent–pixel mapping is no longer enforced to be spatially local, and applying CAA propagates information across unrelated regions, leading to degraded multi-view consistency and reduced overall performance, as observed in Tab.~\ref{tab:ablation}.

\begin{table}[t!]
    \centering
    \renewcommand{\arraystretch}{1.10}
    \begin{tabular}{lccc}
        \toprule
        \textbf{Crop range} &
        $\mathrm{FID}^{\text{CLIP}}_{\text{shaded}}$ &
        $\mathrm{FID}^{\text{CLIP}}_{\text{albedo}}$ &
        c-PSNR \\
        \midrule
        $[3,32]$   & \textbf{3.083} & \textbf{4.599} & \textbf{21.934} \\
        $[24,48]$  & 3.208 & 5.027 & 20.571 \\
        $[32,64]$  & 3.339 & 4.719 & 20.980 \\
        \bottomrule
    \end{tabular}
    \caption{\textbf{Crop-range ablation for locality regularization.} We vary the crop-size range used in $\mathcal{L}_{\text{local}}$ on a $64 \times 64$ latent grid. Smaller crop ranges yield the best overall performance, while all cropped settings outperform training without cropping.}
    \label{tab:locality_ablation}
\end{table}

We further study the sensitivity of locality regularization to the crop size used during training. In our default setting on a $64 \times 64$ latent grid, cropping is applied with 50\% probability, using a square crop whose size is uniformly sampled from $[3,32]$ at a random location. In Tab.~\ref{tab:locality_ablation}, we compare this setting with larger crop ranges, $[24,48]$ and $[32,64]$. All crop ranges outperform training without cropping, while the default range $[3,32]$ achieves the best overall performance. This suggests that smaller patches provide a stronger and more robust locality prior, which is well aligned with CAA operating on local correspondence windows.

\begin{table}[t]
    \centering
    \renewcommand{\arraystretch}{1.15}
    \begin{tabular}{ccc}
        \toprule
        \(\mathrm{SWD}(\mathbf{z}, \mathbf{z}_{\mathrm{a}})\) &
        \(\mathrm{SWD}(\mathbf{z}, \mathbf{z}_{\mathrm{rm}})\) &
        \(\mathrm{SWD}(\mathbf{z}, \mathbf{z}_{\mathrm{ours}})\) \\
        \midrule
        0.539 & 0.897 & 0.560 \\
        \bottomrule
    \end{tabular}
    \caption{\textbf{Sliced Wasserstein distances to the pretrained diffusion prior.} Lower values indicate that the encoded latents are better aligned with the pretrained latent space.}
    \label{tab:sw2_2x3}
\end{table}

\subsection{Analysis of Latent Distribution Mismatch}
\label{subsec:latent_mismatch}
As discussed in Sec.~\ref{subsec:mcvae}, a key challenge in adapting an RGB-pretrained diffusion model to PBR texture generation is the latent distribution mismatch caused by additional material channels. To mitigate this, \Ours{} regularizes the learned latents toward the pretrained latent space.

Tab.~\ref{tab:sw2_2x3} reports sliced Wasserstein distances between the pretrained diffusion prior and encoded PBR latents. We estimate the SD3.5-medium latent prior from 5K samples generated using the text prompts paired with our PBR dataset, and compare it with the latents obtained by encoding the corresponding material maps using \matvae{} and the pretrained VAE.

The pretrained VAE latent of albedo, $\mathbf{z}_{\mathrm{a}}$, remains relatively close to the prior, $\mathbf{z}$, while the latent of roughness+metallic, $\mathbf{z}_{\mathrm{rm}}$, exhibits a larger discrepancy. This supports our claim that additional PBR channels induce a mismatch with the pretrained diffusion latent space. In contrast, \matvae{} reduces this mismatch substantially, suggesting that $\mathcal{L}_{\mathrm{reg}}$ helps preserve alignment with the pretrained latent space while incorporating material information.

\section{Implementation Details}
\label{sec:impl_details}
In this section, we present implementation details for \matvae{} (Sec.~\ref{subsec:matvae_impl}) and \Ours{} (Sec.~\ref{subsec:matlat_impl}), along with the data preprocessing pipeline (Sec.~\ref{subsec:data_preprocess}). 

\subsection{\matvae{}}
\label{subsec:matvae_impl}
\paragraph{Architecture.}
Our \textsc{MatVAE} is initialized from the pretrained VAE of \texttt{Stable Diffusion 3.5-Medium}. 
We freeze the original encoder \(\mathcal{E}_{\text{pre}}\) and decoder \(\mathcal{D}_{\text{pre}}\), and introduce an offset encoder \(\mathcal{E}_{\text{res}}\) to adapt the latent space to 5-channel PBR inputs.

The offset encoder $\mathcal{E}_{\text{res}}$ shares the same architecture as $\mathcal{E}_{\text{pre}}$ except for the first convolution layer, which is modified to accept 5-channel inputs $[\mathbf{a}, \mathbf{r}, \mathbf{m}]$. All intermediate layers of $\mathcal{E}_{\text{res}}$ are initialized from the corresponding weights of $\mathcal{E}_{\text{pre}}$, while the output layer is zero-initialized such that $\boldsymbol{\mu}_{\text{res}} = \mathbf{0}$ and $\boldsymbol{\sigma}_{\text{res}} = \mathbf{1}$ at initialization. The decoder $\mathcal{D}$ extends the final convolution of $\mathcal{D}_{\text{pre}}$ to output 5 channels; all other layers are copied from the pretrained decoder.

\paragraph{Training Configurations.}
We train \textsc{MatVAE} with the loss function in Eq.~\ref{eq:matvae_loss}. 
We use the Adam optimizer with learning rate $3\times10^{-5}$ and batch size $8$, training for 200k iterations on $8$ NVIDIA RTX Pro$6000$ GPUs for $60$ hours. 
The loss weights are set to $\lambda_{\text{local}}=3$, $\lambda_{\text{KL}}=10^{-6}$, $\lambda_{\text{disc}}=0.02$, and $\lambda_{\text{reg}}=3\times10^{-9}$. 
For locality regularization, we randomly crop a square region covering $0.2\%$–$25\%$ of the image area with probability \(0.5\).

\subsection{Diffusion Model Fine-Tuning}
\label{subsec:matlat_impl}
Our diffusion model is built on \textsc{Stable Diffusion 3.5-Medium} with MMDiT~\cite{esser2024scaling} backbone. 
For each joint attention layer, we introduce a parallel correspondence-aware attention (CAA) branch that attends over geometrically corresponding pixels across views, using the precomputed 3D correspondences described in Sec.~\ref{subsec:mvdit}. 
The CAA branch uses the same base projection weights as the original joint attention layers, while introducing additional LoRA layers~\cite{hu2022lora} with rank $32$. The CAA output is added residually to the original attention output. 

Additionally, to align the generated images with the input geometry, we follow previous works~\cite{bensadoun2024meta, Huang:2025mvadapter, He:2025materialmvp} and condition the diffusion model on rendered position and normal maps from the corresponding camera view. 
These geometric features are concatenated with the noisy latent to form the diffusion input.

\paragraph{Training Configurations.}
We optimize \Ours{} using the Conditional Flow Matching~\cite{Lipman:2023flowmatching} objective defined in Eq.~\ref{eq:matlat_loss}, where timesteps are sampled from the same logit-normal distribution as in SD3~\cite{esser2024scaling}. 
We use the Adam optimizer with learning rate $5\times10^{-5}$ and batch size $4$, training for $20$k iterations on $8$ NVIDIA RTX Pro$6000$ GPUs, which takes about $24$ hours in total.

\paragraph{Inference.}
At inference time, we generate multi-view images from \(N = 6\) canonical views (front, back, left, right, top, and bottom) using the Euler sampler with $30$ steps and a Classifier-Free Guidance~\cite{ho2022classifier} scale of $4.0$. 
Given the generated PBR latent samples, we first decode them into 5-channel material images and then convert the multi-view PBR outputs into a final UV texture map following the pipeline of MVAdapter~\cite{Huang:2025mvadapter}: we upscale the images, unproject them into UV space using the given camera poses and mesh, and finally perform inpainting in UV space to fill occluded regions. 

\paragraph{Baseline Implementations.}
\label{subsec:baseline_impl}
For MeshGen~\cite{Chen:2025meshgen} and MaterialMVP~\cite{He:2025materialmvp}, which operate in an image-conditional setting, we use \textsc{Stable Diffusion 2-Depth} to generate the reference images used as conditional inputs. 
All other baselines are evaluated using their official implementations with default configurations.

\newlength{\figw}
\newlength{\envsize}
\newlength{\ressize}
\setlength{\figw}{\textwidth}       %
\setlength{\envsize}{0.06\figw}     %
\setlength{\ressize}{0.14\figw}     %
\setlength{\fboxrule}{0.5pt}
\setlength{\fboxsep}{0pt}

\begin{figure*}[t!]
  \centering
  \setlength{\tabcolsep}{0pt}
  \renewcommand{\arraystretch}{1.0}
  \begin{tabular}{@{}*{8}{>{\centering\arraybackslash}m{0.125\linewidth}}@{}}
    \includegraphics[width=\envsize,height=\envsize,keepaspectratio,valign=m]{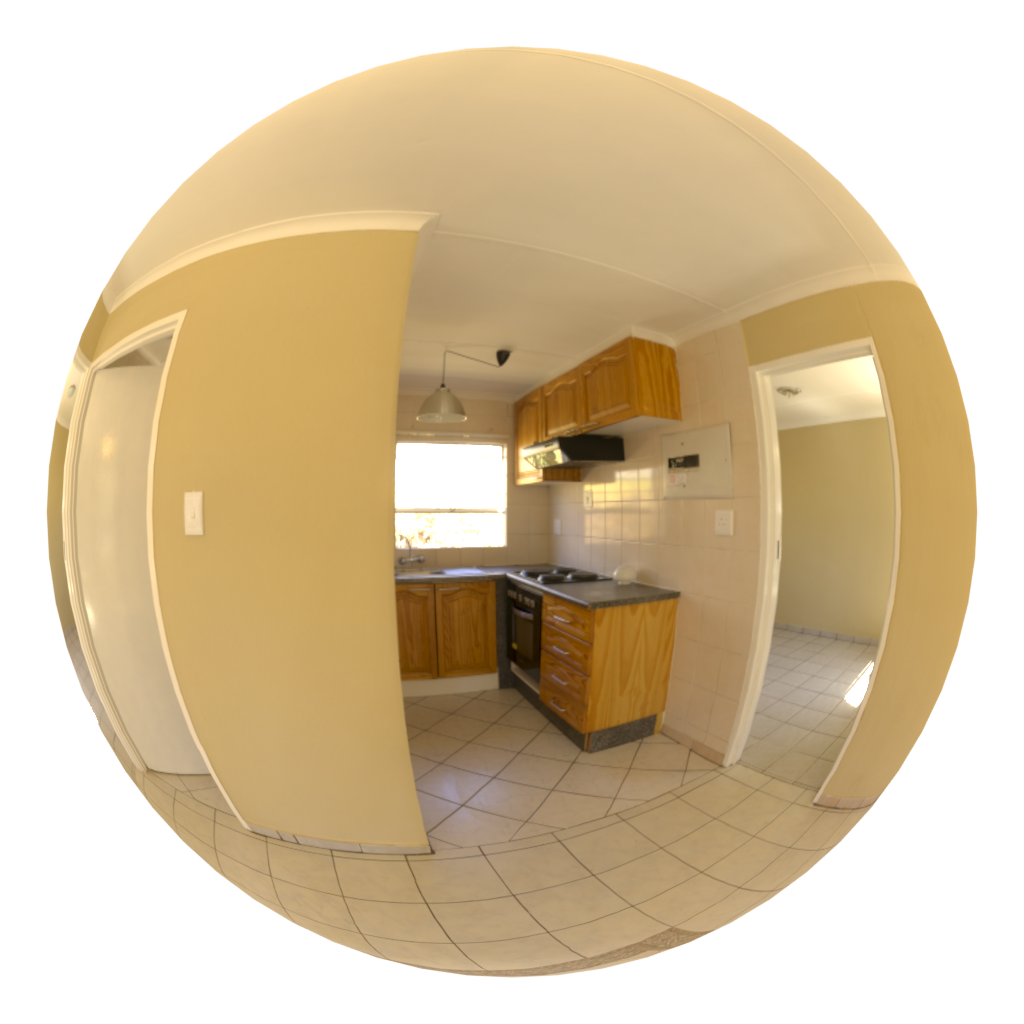} &
    \includegraphics[width=\envsize,height=\envsize,keepaspectratio,valign=m]{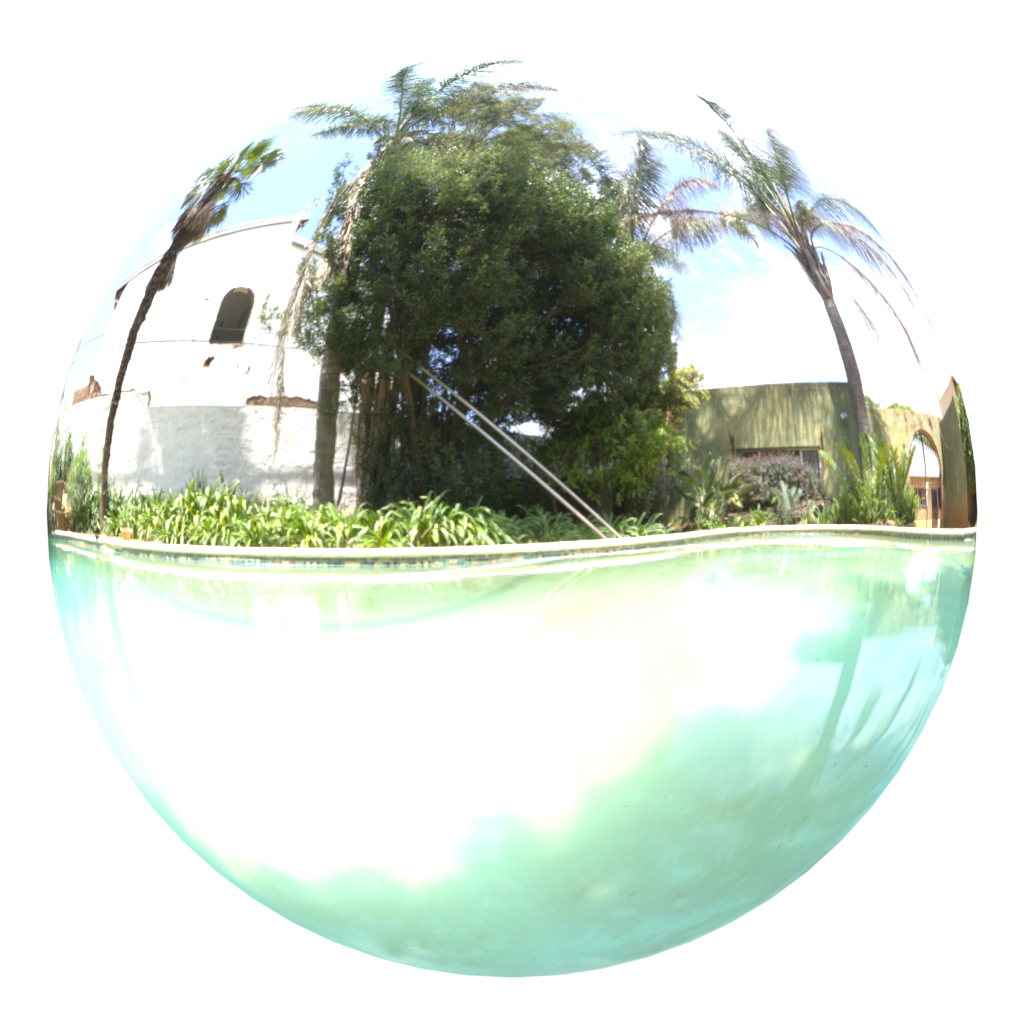} &
    \includegraphics[width=\envsize,height=\envsize,keepaspectratio,valign=m]{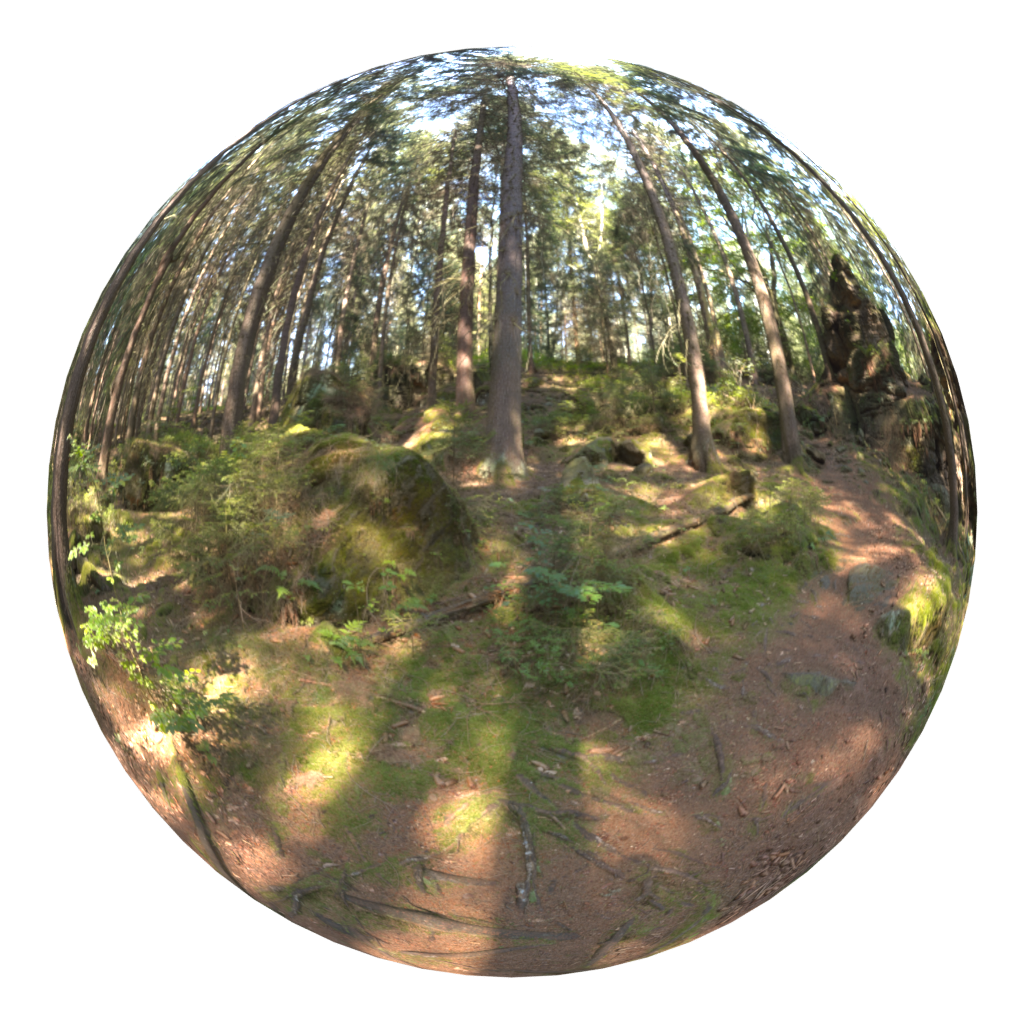} &
    \includegraphics[width=\envsize,height=\envsize,keepaspectratio,valign=m]{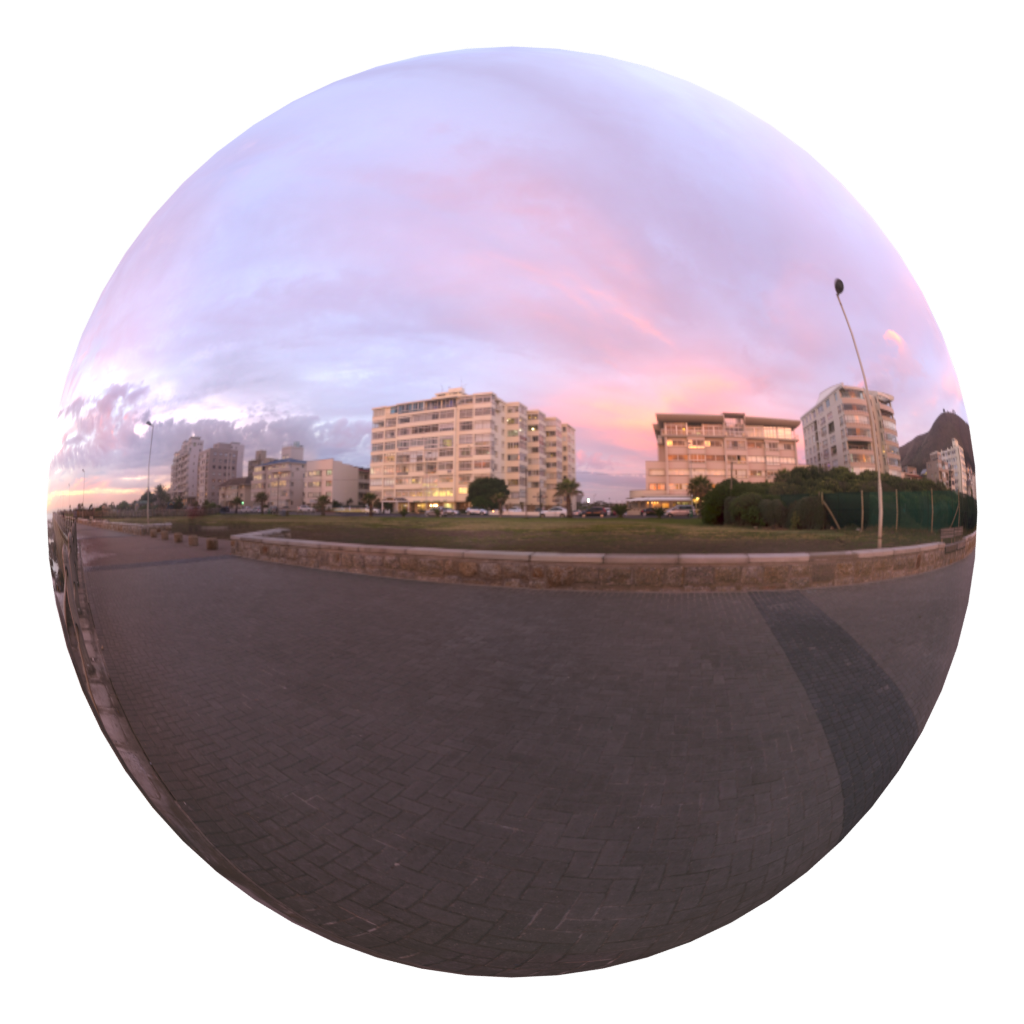} &
    \includegraphics[width=\envsize,height=\envsize,keepaspectratio,valign=m]{figures/relighting/envmaps/hdr1.png} &
    \includegraphics[width=\envsize,height=\envsize,keepaspectratio,valign=m]{figures/relighting/envmaps/hdr3.png} &
    \includegraphics[width=\envsize,height=\envsize,keepaspectratio,valign=m]{figures/relighting/envmaps/hdr4.png} &
    \includegraphics[width=\envsize,height=\envsize,keepaspectratio,valign=m]{figures/relighting/envmaps/hdr5.png} \\
    \makebox[\ressize][c]{\includegraphics[width=\ressize,height=\ressize,keepaspectratio,valign=m]{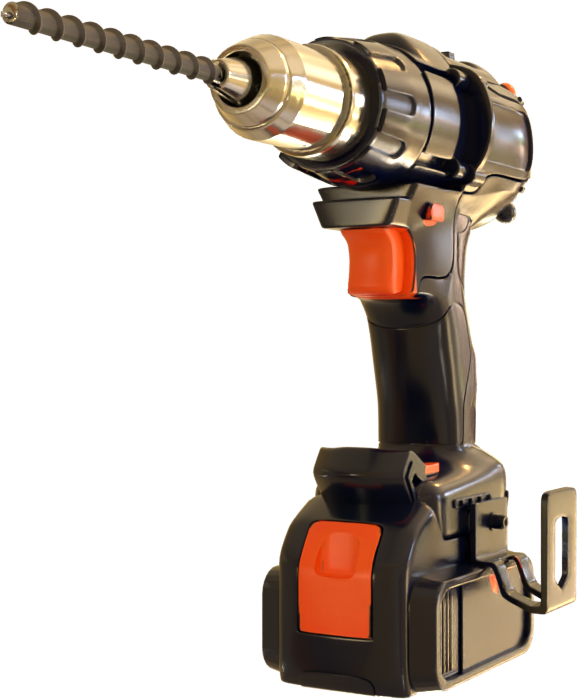}} &
    \makebox[\ressize][c]{\includegraphics[width=\ressize,height=\ressize,keepaspectratio,valign=m]{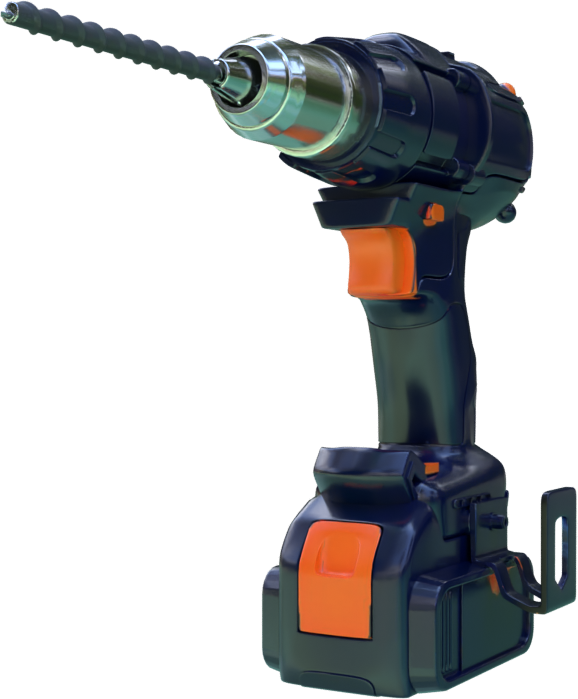}} &
    \makebox[\ressize][c]{\includegraphics[width=\ressize,height=\ressize,keepaspectratio,valign=m]{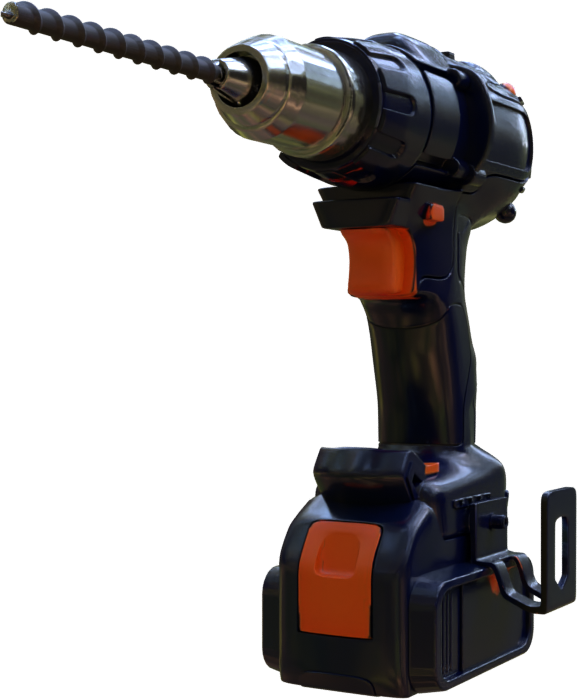}} &
    \makebox[\ressize][c]{\includegraphics[width=\ressize,height=\ressize,keepaspectratio,valign=m]{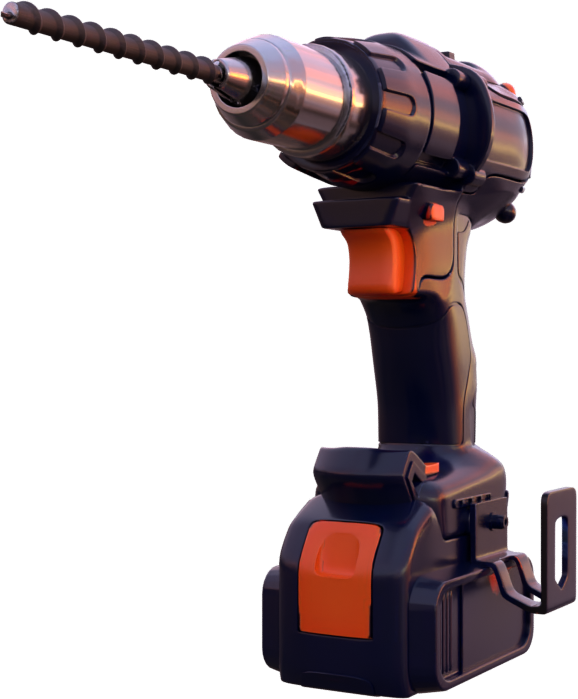}} &
    \makebox[\ressize][c]{\includegraphics[width=\ressize,height=\ressize,keepaspectratio,valign=m]{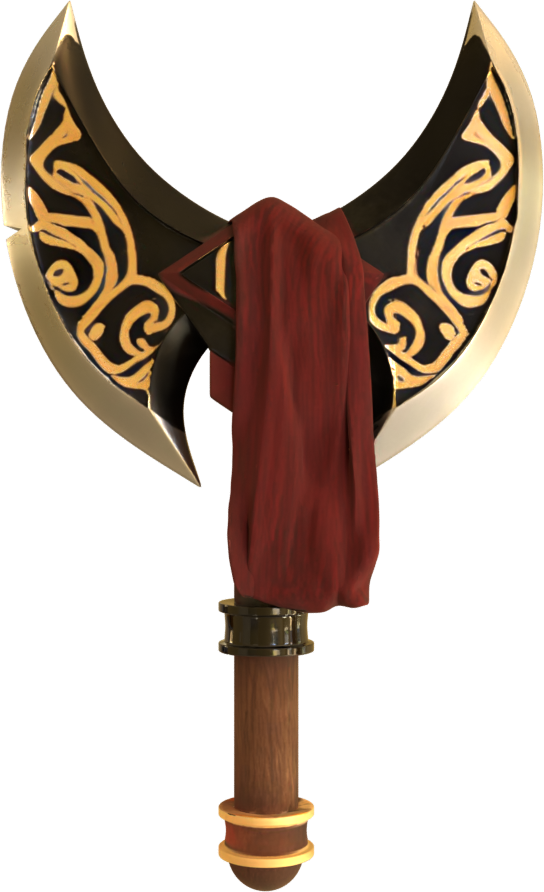}} &
    \makebox[\ressize][c]{\includegraphics[width=\ressize,height=\ressize,keepaspectratio,valign=m]{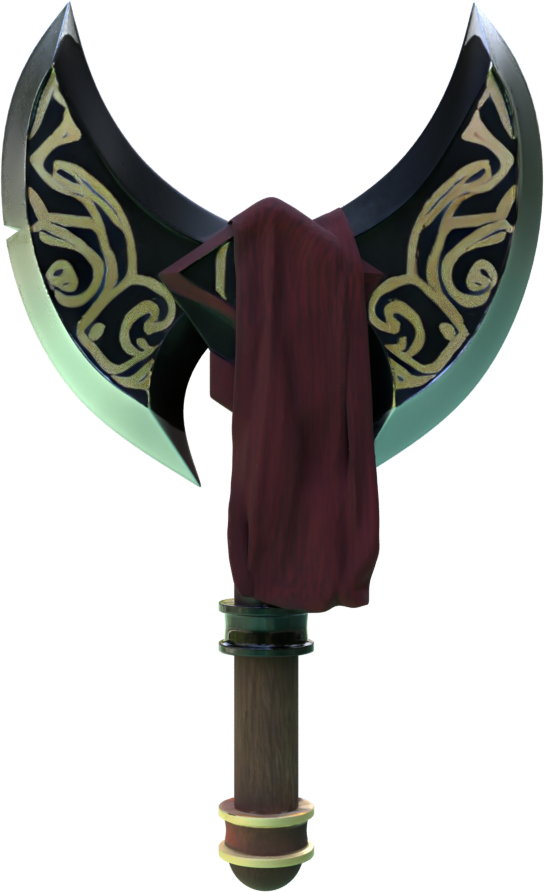}} &
    \makebox[\ressize][c]{\includegraphics[width=\ressize,height=\ressize,keepaspectratio,valign=m]{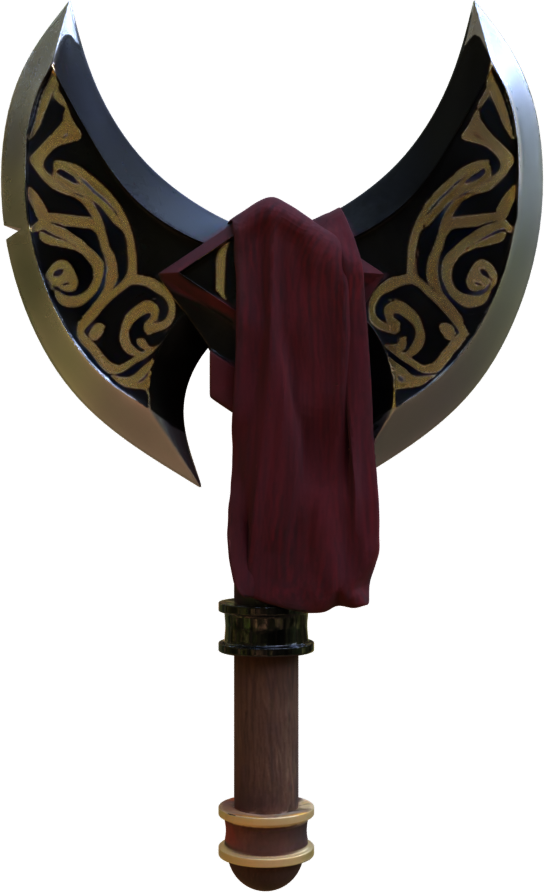}} &
    \makebox[\ressize][c]{\includegraphics[width=\ressize,height=\ressize,keepaspectratio,valign=m]{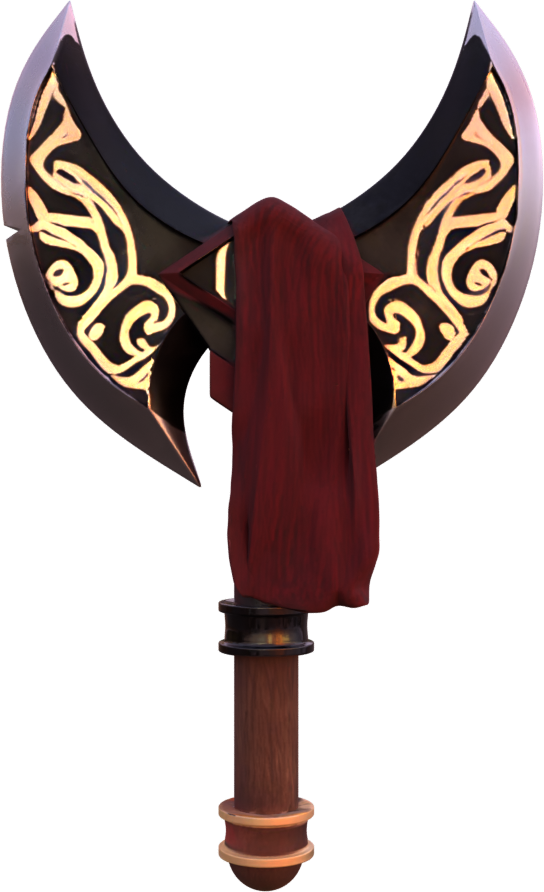}} \\
  \end{tabular}
  \caption{\textbf{Relighting results.} Example objects generated by \Ours{} and rendered under four different environment maps.}
  \label{fig:supp_relighting}
\end{figure*}

\subsection{Data Processing}
\label{subsec:data_preprocess}
We curate $40{,}851$ meshes with PBR textures from Objaverse-XL~\cite{Deitke:2023objaversexl}, holding out $128$ meshes for evaluation.
For each mesh, we render material images from $26$ fixed camera views surrounding the object. 
We then render albedo, roughness, and metallic images, along with normal and position maps for conditioning, from each view.
During diffusion training, we randomly sample $6$ of these $26$ views per iteration to construct the multi-view training batch.
Additionally, for text conditioning, we use captions from Bootstrap3D~\cite{Sun:2025bootstrap3d} when available; otherwise, we use captions from Cap3D~\cite{Luo:2023cap3d}.

During evaluation, we render the PBR-textured assets under environment lighting using $785$ HDR environment maps from Poly Haven. 
For each image, we then randomly sample an environment map and apply the same rendering setup to both the generated and ground-truth images to ensure a fair comparison.

\section{Additional Qualitative Results}
\label{sec:quali_res}
In this section, we present representative qualitative examples for the ablation studies (Sec.~\ref{subsec:supp_ablation}), baseline comparisons (Sec.~\ref{subsec:supp_baseline}), and generated textures of \Ours{} (Sec.~\ref{subsec:supp_showcase}). \textbf{Please refer to the project page for more results and video demonstrations:} \url{https://matlat-proj.github.io/}

\subsection{Additional Qualitative Results for Ablation Studies}
\label{subsec:supp_ablation}
Fig.~\ref{fig:ablations} extends the ablation study in Sec.~\ref{subsec:ablation} with additional qualitative examples. 
We observe that Frozen VAE produces unrealistic material appearance, as evidenced by the overly shiny metallic shoe surface. 
Additionally, Res. Pred. + $\mathcal{L}_{\text{id}}$ and Direct Pred. + $\mathcal{L}_{\text{reg}}$ miss fine details such as \textit{metal} buckles, whereas \Ours{} generates PBR textures with correct and realistic material appearances. 

\subsection{Additional Qualitative Comparisons with Previous Methods}
\label{subsec:supp_baseline}
Fig.~\ref{fig:baselines} presents qualitative comparisons against representative baselines, including models trained from scratch (MeshGen~\cite{Chen:2025meshgen}, TexGaussian~\cite{Xiong:2025texgaussian}), SDS-based optimization methods (Paint-it~\cite{Youwang:2024paintit}, DreamMat~\cite{Zhang:2024dreammat}, FlashTex~\cite{Deng:2024flashtex}), and prior multi-view diffusion models (MaterialAnything~\cite{Huang:2024materialanything}, MaterialMVP~\cite{He:2025materialmvp}).
Models trained from scratch often produce textures with suboptimal quality and exhibit weak alignment with the input text prompts. 
Additionally, SDS-based methods lack fine details and tend to generate textures with oversaturated colors. 
In contrast, our method \Ours{} generates high-quality textures with realistic material properties.

\subsection{Additional Qualitative Results of \Ours{}}
\label{subsec:supp_showcase}
Extending the results in Fig.~\ref{fig:main_quali}, we present additional PBR textures generated by our method \Ours{} with diverse prompts, meshes, and environment maps in Fig.~\ref{fig:ours}. 
Note that our method generates physically plausible material properties that are well aligned with both the object characteristics and the provided text prompt. 
These results demonstrate that our pipeline exhibits strong generalization across diverse object categories and material types.

Fig.~\ref{fig:supp_relighting} further presents relighting examples, where the same object is rendered under four different environment maps. The consistent appearance changes across lighting conditions demonstrate that \Ours{} produces coherent PBR textures with plausible illumination-dependent effects.

\clearpage
\newpage

\begin{table*}[h]
\centering
\begin{tabularx}{\linewidth}{X}
    \begin{minipage}{\linewidth}
        \centering
        \includegraphics[width=\linewidth]{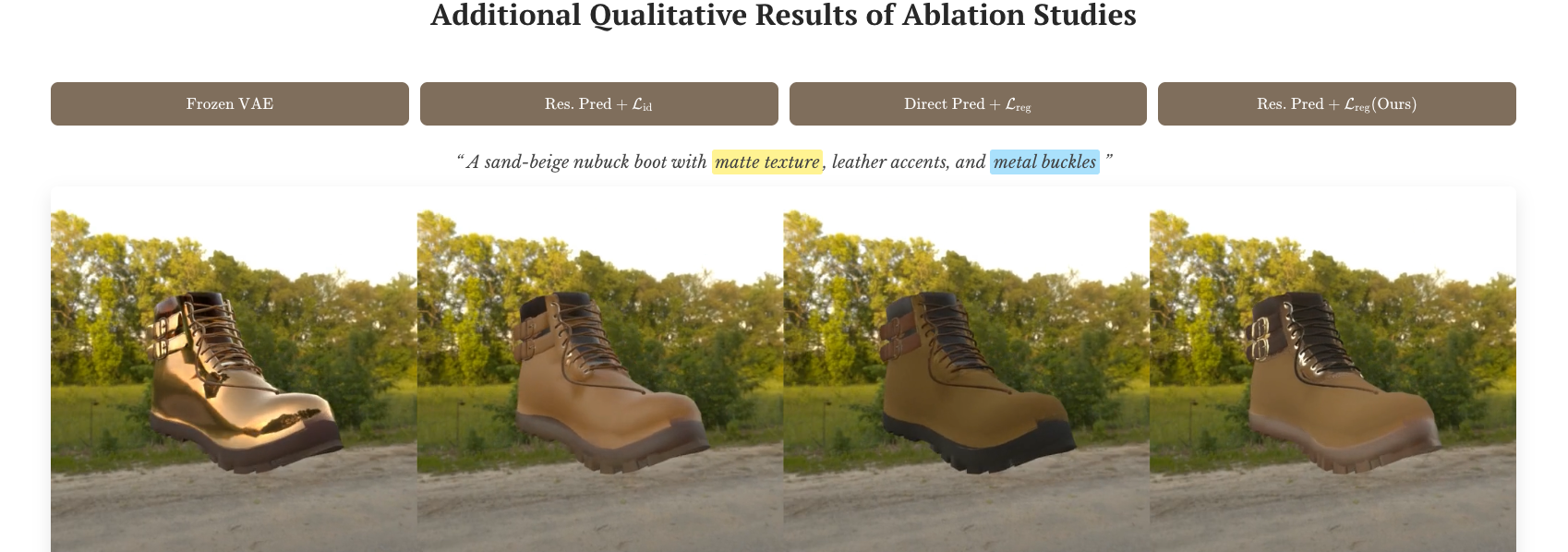}
        \captionof{figure}{
            \textbf{Additional Qualitative Results of Ablation Studies.}
            Extended qualitative results of the ablation studies presented in Fig.~\ref{fig:ablation_qualitative} \refofpaper{}: Frozen VAE, Res. Pred. + $\mathcal{L}_\text{id}$, Direct Pred. + $\mathcal{L}_\text{reg}$, and Res. Pred. + $\mathcal{L}_\text{reg}$ (Ours). 
            \textbf{Best viewed in video.}
        }
        \label{fig:ablations}
    \end{minipage}
    \\ \vspace{0.5em}

    \begin{minipage}{\linewidth}
        \centering
        \includegraphics[width=\linewidth]{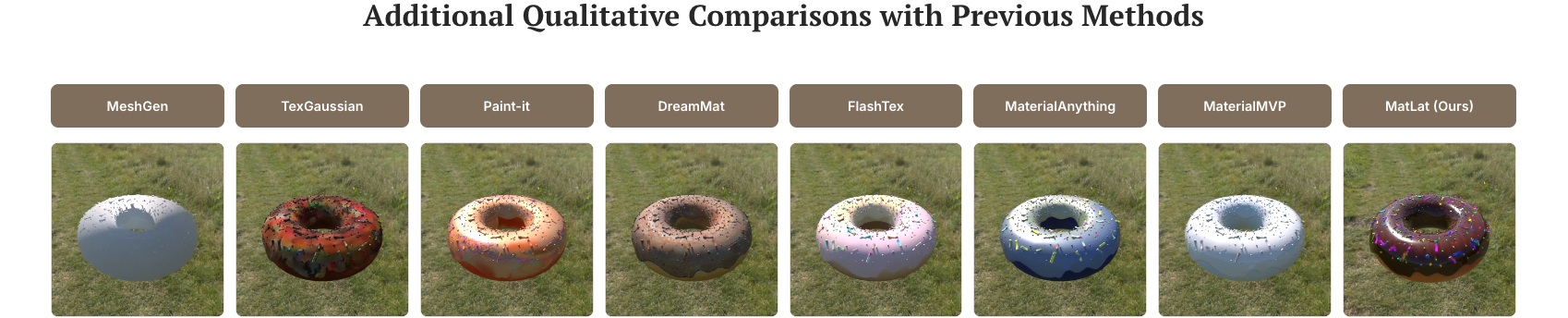}
        \captionof{figure}{
            \textbf{Additional Qualitative Comparisons with Previous Methods.}
            Each column shows the shaded output rendered under identical lighting conditions. Our method produces high-quality, physically plausible materials with superior text-visual alignment and detail preservation. 
            \textbf{Best viewed in video.}
        }
        \label{fig:baselines}
    \end{minipage}
    \\ \vspace{0.5em}

    \begin{minipage}{\linewidth}
        \centering
        \includegraphics[width=\linewidth]{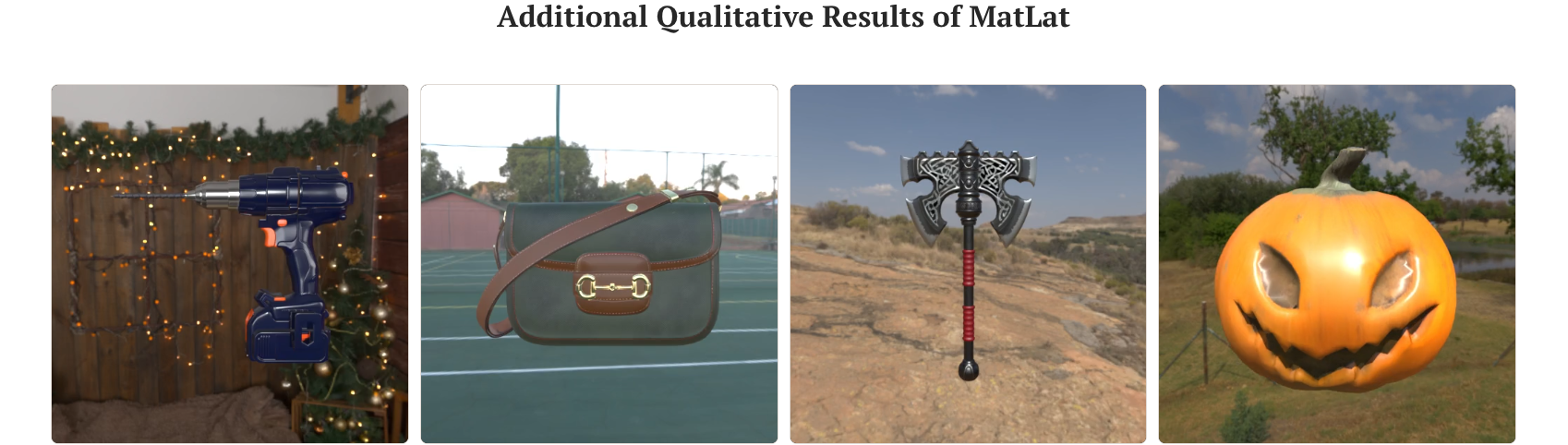}
        \captionof{figure}{
            \textbf{Additional Qualitative Results of \Ours{}.}
            Gallery of assets textured by \Ours{} for various text prompts and environment maps. 
            \textbf{Best viewed in video.}
        }
        \label{fig:ours}
    \end{minipage}
    \\
    \vspace{3em}
    \large{
    }

    \large
    
    \vspace{1.5em}

\end{tabularx}
\end{table*}

\end{document}